\documentclass[lettersize,journal]{IEEEtran}
\usepackage{amsmath,amsfonts}
\usepackage{array}
\usepackage{color}
\usepackage{textcomp}
\usepackage{stfloats}
\usepackage{url}
\usepackage{graphicx}
\usepackage{subfigure}
\usepackage{cite}
\usepackage[ruled]{algorithm2e} 
\usepackage{diagbox} 
\usepackage{booktabs}  
\usepackage{multirow}  
\usepackage{makecell}  
\usepackage{cite}
\usepackage{url} 
\begin{document}

\title{Semantic Geometric Fusion Multi-object Tracking and Lidar Odometry in Dynamic Environment}

\author{Tingchen Ma and Yongsheng Ou*
\thanks{This work was supported by the National Key Research and Development Program of China under Grant 2018AAA0103001; in part by the National Natural Science Foundation of China (Grants No. U1813208, 62173319, 62063006); in part by the Guangdong Basic and Applied Basic Research Foundation (2020B1515120054); in part by the Shenzhen Fundamental Research Program (JCYJ20200109115610172).}
\thanks{* Corresponding author, e-mail: ys.ou@siat.ac.cn.}
\thanks{Tingchen Ma is with the Shenzhen Institute of Advanced Technology, Chinese Academy of Sciences, Shenzhen 518055, China, and also with the Shenzhen College of Advanced Technology, University of Chinese Academy of Sciences, Shenzhen 518055, China.}
\thanks{Yongsheng Ou is with the Key Laboratory of Human-Machine-Intelligence-Synergy Systems, Shenzhen Institute of Advanced Technology, Chinese Academy of Sciences, Shenzhen 518055, China, and also with the Guangdong Provincial Key Laboratory of Robotics and Intelligent System, Shenzhen Institute of Advanced Technology, Chinese Academy of Sciences, Shenzhen 518055, China}
}

\markboth{Journal of \LaTeX\ Class Files,~Vol.~14, No.~8, August~2021}%
{Shell \MakeLowercase{\textit{et al.}}: A Sample Article Using IEEEtran.cls for IEEE Journals}


\maketitle
\pagestyle{empty}  
\thispagestyle{empty} 

\begin{abstract}
The SLAM system based on static scene assumption will introduce huge estimation errors when moving objects appear in the field of view. This paper proposes a novel multi-object dynamic lidar odometry (MLO) based on semantic object detection technology to solve this problem. The MLO system can provide reliable localization of robot and semantic objects and build long-term static maps in complex dynamic scenes. For ego-motion estimation, we use the environment features that take semantic and geometric consistency constraints into account in the extraction process. The filtering features are robust to semantic movable and unknown dynamic objects. At the same time, a least square estimator using the semantic bounding box and object point cloud is proposed to achieve accurate and stable multi-object tracking between frames. In the mapping module, we further realize dynamic semantic object detection based on the absolute trajectory tracking list (ATTL). Then, static semantic objects and environmental features can be used to eliminate accumulated localization errors and build pure static maps. Experiments on public KITTI data sets show that the proposed system can achieve more accurate and robust tracking of the object and better real-time localization accuracy in complex scenes compared with existing technologies.
\end{abstract}


\section{Introduction}

Simultaneous localization and mapping (SLAM) \cite{zhang2014loam, shan2018lego} technology using 3D lidar is widely used in industry due to its good scene robustness (lighting, low texture). Meanwhile, the calculation result provided by the SLAM system is also essential for robot control and task planning. Currently, most SLAM systems \cite{mur2015orb, behley2018efficient} are built based on rigid assumptions, which will introduce huge estimation errors when there are many moving objects in the scene. Therefore, it is necessary to take reasonable measures to improve the performance of robot navigation systems in real scenes.

In recent years, some studies have considered detecting dynamic objects through spatial geometric constraints \cite{deschaud2018imls, moosmann2010motion, park2022nonparametric}, then the SLAM system can avoid the interference of abnormal data association on ego localization. On the other hand, the learning-based semantic detection method is also an excellent way to process dynamic objects. For example, \cite{ruchti2018mapping} and \cite{pfreundschuh2021dynamic} directly identify moving 3d points in a single scan through the trained network model. \cite{chen2021psf} and \cite{chen2019suma++} use semantic segmentation network \cite{milioto2019rangenet++} to obtain point-level semantic labels in the lidar scan, and realize 3D map construction with scene semantics (roads, buildings).

In fact, both geometry-based and semantic-based detection and elimination schemes can achieve reliable robot localization in the dynamic scene. However, for some practical applications such as planning tasks in autonomous driving and human-computer interaction in AR/VR, robots further need to stably track and represent semantic objects (cars and people) in the map coordinate. We propose a least square estimator for the above problems that uses semantic and geometric information for reliable multi-object tracking (SGF-MOT). Combined with the estimation results of ego odometry, the absolute localization for each object can be calculated. In the mapping module, the absolute trajectory tracking list of objects is used for dynamic semantic object detection. Then, static object and environment features can be used to eliminate the accumulated error of ego and object odometry. Finally, we realize the construction of a 4D scene map.

The main contributions of this paper are as follows: 

\begin{itemize}
\item[$\bullet$] A complete multi-object lidar odometry system (MLO) can provide the absolute localization of both robot and semantic objects and build a 4D scene map (robot, semantic object localization and long-term static map). 
\end{itemize}

\begin{itemize}
\item[$\bullet$] A least-square estimator considering the bounding box plane feature and geometric point cloud distribution is used for object state updating in the multi-object tracking module. The semantic bounding boxes can ensure the correct optimization direction. The relative motion model eliminates the point cloud distortion caused by robot and object motion. Meanwhile, the direct measurement point cloud can improve the estimation accuracy. 
\end{itemize}

\begin{itemize}
\item[$\bullet$] A dynamic object detection method based on the absolute trajectory tracking list, which can identify slow-moving semantic objects.
\end{itemize}

The rest of this paper is structured as follows. First, in section 2, we discuss related works. Then, the proposed MLO system are described in section 3. The experimental setup, results, and discussion will be presented in section 4. Finally, section 5 is the conclusion.

\begin{figure}[t]
\centering
\includegraphics[scale=0.30]{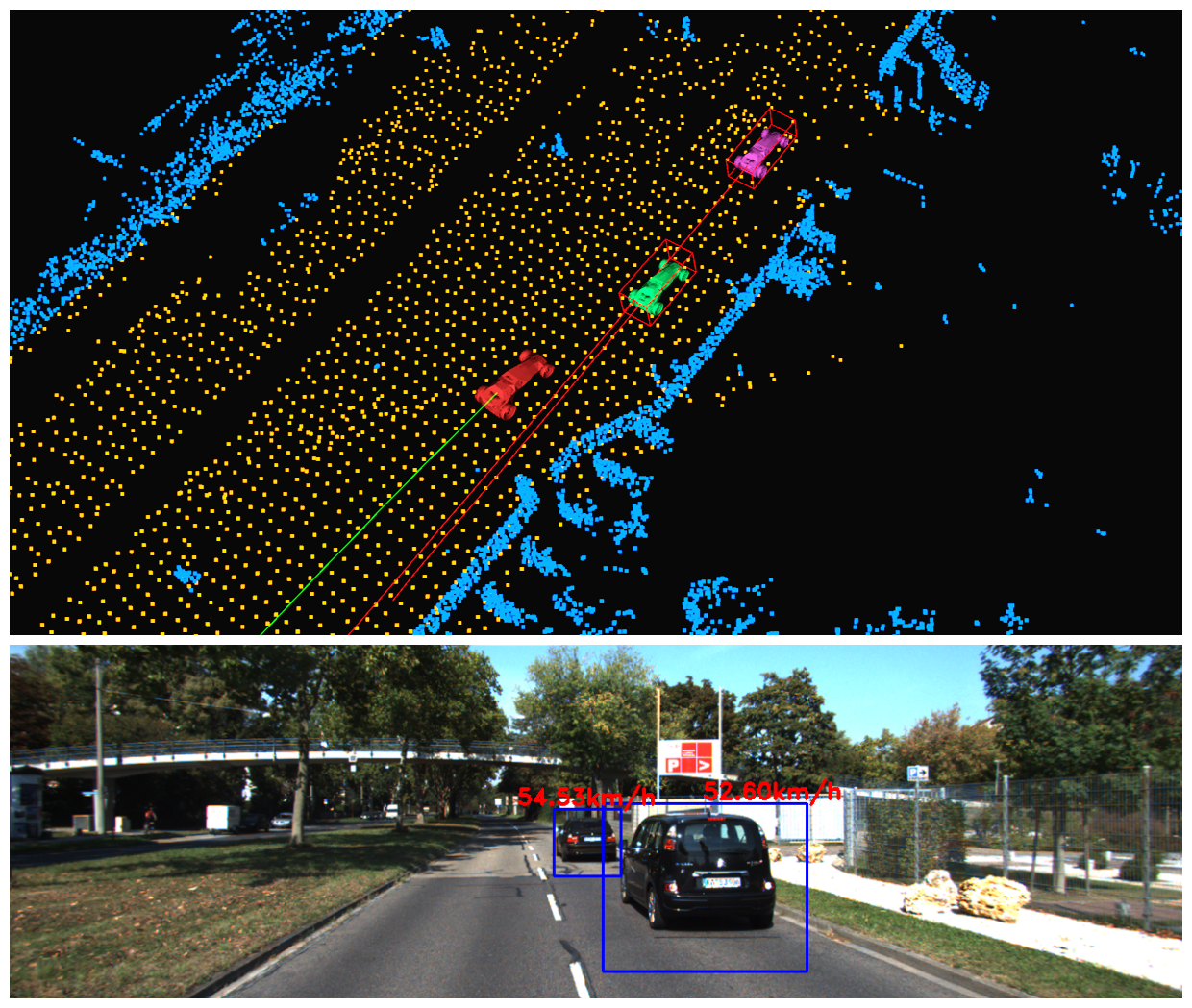}
\caption{{\bf{Qualitative results for sequence 0926-0013 in the KITTI-raw dataset.}} The top half of the picture shows the results in rviz. The red car is a robot equipped with lidar, and green line is the ego motion path. The red lines are the tracked object motion paths. The bottom half (for visualization only) shows the projected 2D bounding box (blue) and corresponding object velocity (red) in the image with the synchronized timestamp. }
\label{fig_framework2}
\end{figure}

\section{Related work}

\subsection{Object detection and tracking}

For the multi-object tracking (MOT) method based on point cloud data, object detection and tracking are the two main steps of most solutions. In the early days, traditional methods focused on extracting objects from the spatial distribution information of point clouds. Then, object tracking is realized by constructing geometric error constraints and corresponding estimator. Moosmann et al. \cite{moosmann2010motion} implemented a general object segmentation method based on the local convexity criterion. The point-to-plane ICP \cite{chen1992object} algorithm is used for MOT. Dewan et al. \cite{dewan2016motion} used RANSAC \cite{fischler1981random} to estimate the motion model sequentially. The point clouds were then associated with the corresponding model by a bayesian approach. By assuming local geometric constancy and regularization of smooth motion field, Ferri et al. \cite{dewan2016rigid} estimated object motion through rigid scene flow, and their method can be suitable for non-rigid motion, such as pedestrians. Sualeh et al. \cite{liu2021dloam} fitted the 3d bounding box from the cluster point cloud and updated the object state based on the Kalman filter \cite{kalman1960new}.

Recently, the learning-based recognition method on point clouds has brought new solutions to the MOT problem. They can bring more stable object recognition and semantic information (such as object bounding boxes). Weng et al. \cite{weng20203d} use Point-RCNN \cite{shi2019pointrcnn} to detect semantic objects in point clouds and track 3D bounding boxes (3D-BB) through the Kalman filter. Kim et al. \cite{kim2021eagermot} used 2D/3D-BB detected in image and point clouds for filter tracking, achieving better tracking accuracy. Wang et al. \cite{wang2021ditnet} and Huang et al. \cite{huang2021joint} designed an end-to-end MOT framework using different deep-learning models to complete object detection and data association tasks.

In summary, a reliable initial guess is indispensable for the optimization-based tracking method using point cloud geometric constraints. When the initial value of the estimator is poor, many point-level data associations may provide the wrong direction at the early stage of the calculation. On the other hand, extracting bounding boxes from the point cloud or learning method for filter-based tracking is relatively stable. However, it abandons the high-precision measured point cloud. In our SGF-MOT module, semantic bounding box plane and geometric point cloud distribution are used simultaneously in a least-square estimator, which achieves a better balance between tracking accuracy and robustness.

\subsection{Dynamic aware 3d lidar SLAM}

At present, 3d lidar SLAM systems are mainly divided into two categories: feature-based and matching-based. For the feature-based method, Zhang et al. \cite{zhang2014loam} and Shan et al. \cite{shan2018lego}  extracted stable line and surface features from the point cloud by scan line smoothness. Then the extracted features are used for localization and mapping. Liu et al. \cite{liu2021balm} introduced the bundle adjustment (BA) \cite{triggs1999bundle} method, which is more commonly used in visual SLAM, to improve the system mapping accuracy. In the matching-based method, Behley et al. \cite{behley2018efficient} used surfel to model the environment and designed a projection data association method for ego localization. Dellenbach et al. \cite{dellenbach2022ct} used the implicit moving least squares surface to represent the map and realized high-precision mapping by considering the continuous time constraints between frames. The above systems are built on the assumption of the rigid scene and cannot work stably in a highly dynamic scene.

For dynamic-aware 3D lidar SLAM technology, some studies use geometric methods such as background model \cite{park2022nonparametric}, local convexity \cite{moosmann2010motion}, and point cloud clustering \cite{deschaud2018imls} to detect and eliminate information that may violate static assumptions. On the other hand, with the development of deep learning technology, some semantic-based lidar SLAM can also process dynamic objects in the field of view. Pfreundschuh et al. \cite{pfreundschuh2021dynamic} designed a method for the automatic labelling of dynamic objects and detected the dynamic 3d points in the point cloud before the implementation of LOAM \cite{zhang2014loam} on the 3D-MinNet network. Sualeh et al. \cite{liu2021dloam} improved the accuracy of dynamic object detection by considering the space-time constraints of sequence frames input during model training. Both \cite{chen2021psf} and \cite{chen2019suma++} using semantic segmentation networks realize semantic scene mapping and detect dynamic objects through point-by-point checks and feature extraction from the object point cloud.

Unlike the above scheme, our system (MLO) integrates the results of robust ego localization and SGF-MOT module, thus achieving accurate and reliable localization of instance objects in the map coordinate. At the same time, the object absolute trajectory tracking list is used to detect dynamic objects in the scene. So we create a 4D scene map that contains both robot, semantic objects position and static environment.

\begin{figure*}[h]
\centering
\includegraphics[scale=0.26]{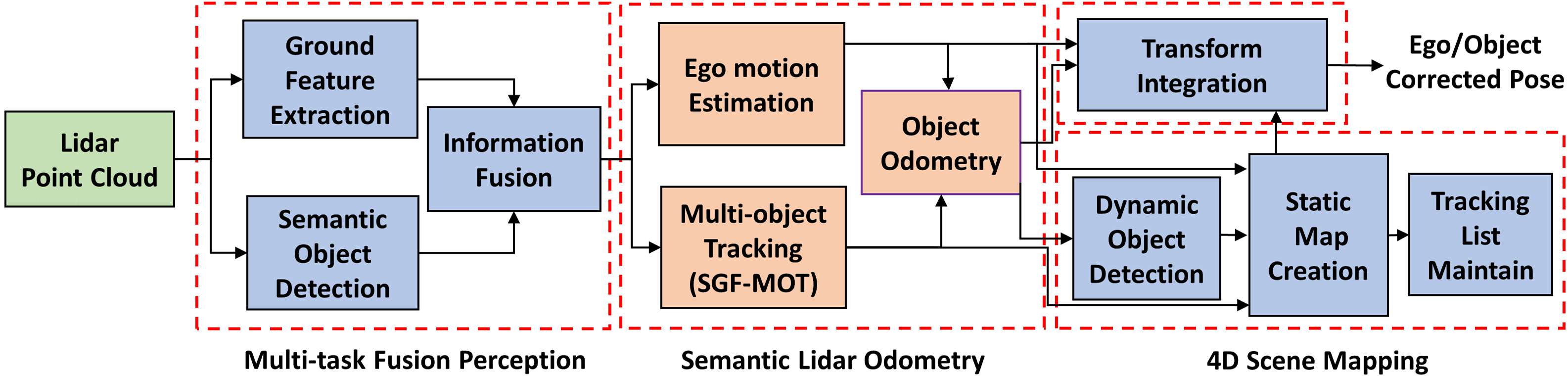}
\caption{{\bf{ The proposed MLO system framework.}} The orange box is the key part of this paper, it includes the semantic-geometric fusion multi-object tracking method (SGF-MOT) which will be discussed in detail. A 4D scene mapping module is also introduced for maintaining the long-term static map and movable object map separately, which can be used to further navigation applications.}
\label{fig_framework4}
\end{figure*}

\section{Method}

The whole MLO system flowchart is shown in Fig.2, which includes three main parts: Multi-task Fusion Perception, Semantic Lidar Odometry, and 4D Scene Mapping. Firstly, the ground and objects are detected and refined by information fusion. Then, the proposed SGF-MOT tracker is integrated into the odometry module to achieve robust and accurate tracking of objects. After dynamic object detection, the mapping module uses both static object and environment features \cite{zhang2014loam} to eliminate accumulated errors.

\subsection{Notation and coordinate frames}

In this paper, we define the notation as follows: The lidar (mechanical) has pose $ {\bf{T}}_{ol}^i \in {\rm{SE}}(3) $ at time $ i $ in odom coordinate $ o $. It can observe movable objects with pose  $ {\bf{T}}_{ot}^i \in {\rm{SE}}(3) $ in odom coordinate $ o $. Meanwhile, the lidar pose in map coordinate $ m $ is $ {\bf{T}}_{ml}^i{\kern 1pt} {\kern 1pt} \in {\rm{SE}}(3) $, and the correction matrix which transforms the object and lidar from odom coordinate $ o $ to map coordinate $ m $ is $ {\bf{T}}_{mo}^{i} \in {\rm{SE}}(3) $.

\subsection{Multi-task fusion perception}

After getting the ROI point clouds ${{\bf{\Phi }}^i} $ of the current frame, we perform ground feature extraction and semantic object detection in parallel to obtain preliminary object point clouds and geometric ground features. Then, the information fusion module can receive more accurate object points segmentation and environment features through a mutual correction step.

\subsubsection{Ground feature extraction}

First, we fit ground parameters in the point cloud using the iterative PCA algorithm proposed in \cite{zermas2017fast} and mark the corresponding points near the ground in point cloud $ {\bf{\Phi }}_g^i $.

The distortion correction time (taking the value range [0,1]) is calculated by the relative offset of each 3D point angle from the 3D point angle at the initial time of the frame. The scan line is uniformly segmented. Through the pointwise inspection of ground labels ({\bf{0:}} background point, {\bf{1:}} ground point) calculated above, it is possible to record whether each segment contains the ground point or background point labels. The segment head and tail point id pairs containing the ground or background points are stored in the ground container $ {{\bf{V}}_g} $ or background container $ {{\bf{V}}_b} $.

Based on the smoothness (readers can refer to \cite{zhang2014loam} for the calculation formula) calculated by segment, ground features $ {\bf{\chi }}_f^i $ can be extracted from the point cloud contained in $ {{\bf{V}}_g} $. The smoothness prior $ {S_{a}} $ used for subsequent background surface feature check is obtained as follows:

$$
{S_{a}} = {{\sum\limits_{o = 1}^N {{c_o}} } \mathord{\left/ {\vphantom {{\sum\limits_{o = 1}^N {{c_o}} } {{N_c}}}} \right. \kern-\nulldelimiterspace} {{N_c}}} \eqno{(1)}
$$

\noindent where $ {c_o} $ is the smoothness of $o$-th ground feature, and $ {N_c} $ is the total number of ground features in current frame. After downsampling all ground points, the candidate ground point cloud $ {\bf{\chi }}_c^i $ is generated.

For point clouds contained in ${{\bf{V}}_b} $, only smoothness is calculated and sorted here. After that, the information fusion module will finely select the background features.

\subsubsection{Semantic object detection} 

The semantic object detection module is to acquire 3D-BB and the foreground object point clouds they enclose. In addition, the 3D-BB will be used to provide optimization constraints for the motion estimation of movable objects. We accomplish the above tasks using 3DSSD \cite{yang20203dssd}, a single-stage, point-based lightweight object detection model that balances detection accuracy and efficiency. Assume that the $k$-th 3D-BB in current detection result is:

$$
{{\bf{\tau }}^{k,i}} = {\left[ {{x^{k,i}},{y^{k,i}},{z^{k,i}},{l^{k,i}},{w^{k,i}},{h^{k,i}},ya{w^{k,i}}} \right]^T} \eqno{(2)}
$$

\noindent where $ \left( {x,y,z} \right) $  represents the object position, $ \left( {l,w,h} \right) $  represents the 3D-BB size, and $ yaw $ is the object direction angle. Each group of object point cloud in 3D-BB is stored as $ {{\bf{\gamma }}^{k,i}} $. Each 3D point is marked with an independent object id in $ {\bf{\Phi }}_s^i $. Object semantic confidences are denoted as $ {\mu ^{k,i}} $.

\begin{figure}[t]
\centering
\includegraphics[scale=0.30]{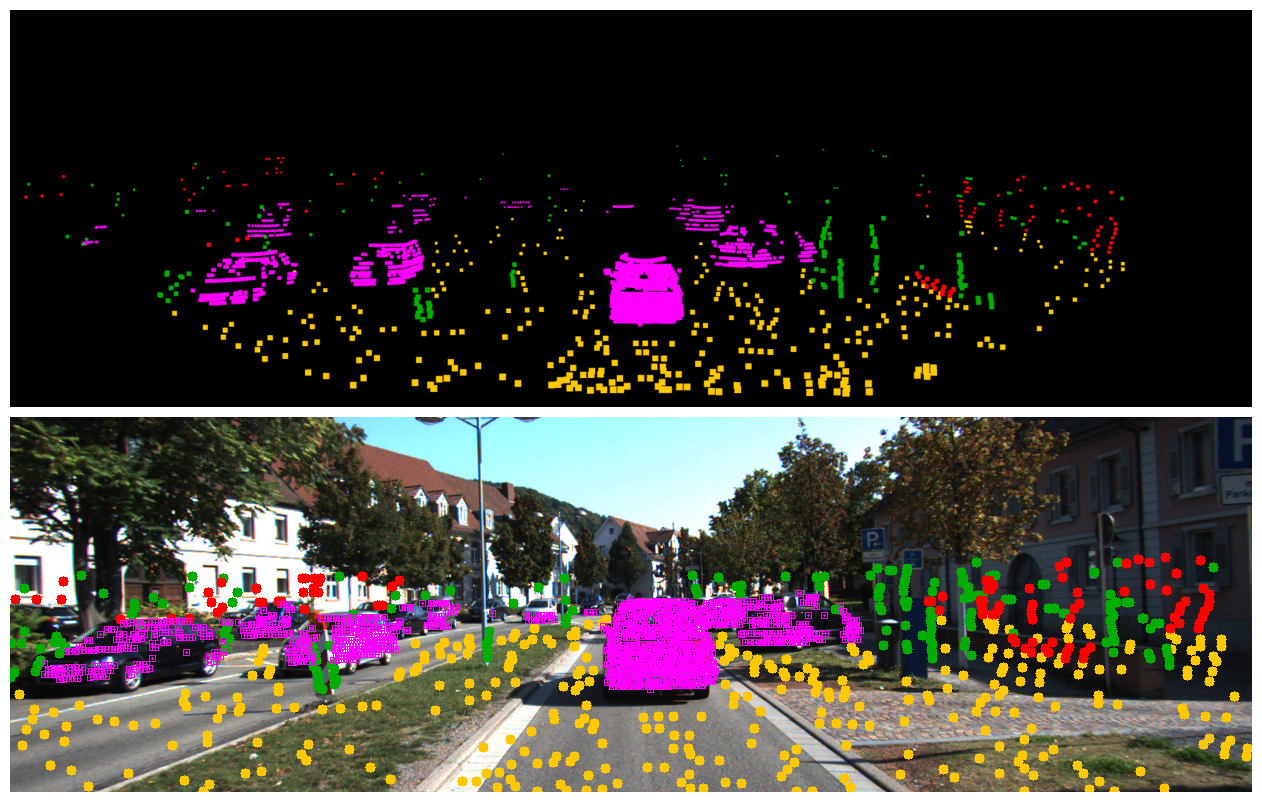}
\caption{{\bf{The results of multi-task fusion perception module for sequence 0926-0056 in the KITTI-raw dataset.}} Pink point clouds represent moving cars, and yellow point clouds represent ground features. Green and red point clouds are background edge and surface features. The image projection results with the same timestamp are used for visualization only. }
\label{fig_framework}
\end{figure}

\subsubsection{Information fusion}

Considering that 3D-BB obtained by semantic detection module usually encloses both object and ground points. Meanwhile, features obtained by ground feature extraction module also contain a small number of points belonging to the object. Therefore, it is necessary to perform mutual correction first to ensure their motion consistency. 

According to the object id marked in $ {\bf{\Phi }}_s^i $, object points in ground feature sets $ {\bf{\chi }}_f^i $ and $ {\bf{\chi }}_c^i $ are eliminated. Similarly, the ground feature in each object point cloud $ {{\bf{\gamma }}^{k,i}} $ can be rejected by labels in $ {\bf{\Phi }}_g^i $. The ambiguous intersection information will not participate in the subsequent SLAM and MOT tasks.

For point clouds contained in ${{\bf{V}}_b} $, we perform two feature extractions based on the computed smoothness: The candidate object edge feature $ {\bf{\gamma }}_c^{k,i} $ is extracted from the point clouds belonging to objects, which will be used in mapping module to eliminate accumulated errors. Then, background edge features $ {\bf{\psi }}_f^i $, surface features $ {\bf{\zeta }}_f^i $, and the corresponding candidate sets $ {\bf{\psi }}_c^i $, $ {\bf{\zeta }}_c^i $ are extracted from the point clouds that do not belong to semantic objects and ground features, and they will participate in both localization and mapping modules.

Finally, we define the edge and surface (including ground) features used for state estimation in this paper. The distance from the point to the edge or surface features in lidar coordinate $ l $ are as follows:

$$
{{\bf{e}}_{{e_k}}} = \frac{{\left| {({}^l{\bf{p}}_{i,k}^e - {}^l{\bf{\bar p}}_{i - 1,m}^e) \times ({}^l{\bf{p}}_{i,k}^e - {}^l{\bf{\bar p}}_{i - 1,n}^e)} \right|}}{{\left| {{}^l{\bf{\bar p}}_{i - 1,m}^e - {}^l{\bf{\bar p}}_{i - 1,n}^e} \right|}} \eqno{(3)}
$$

$$
{{\bf{e}}_{{s_k}}} = \frac{{\left| \begin{array}{c}
\left( {{}^l{\bf{p}}_{i,k}^s - {}^l{\bf{\bar p}}_{i - 1,m}^s} \right)\\
({}^l{\bf{\bar p}}_{i - 1,m}^s - {}^l{\bf{\bar p}}_{i - 1,n}^s) \times ({}^l{\bf{\bar p}}_{i - 1,m}^s - {}^l{\bf{\bar p}}_{i - 1,o}^s)
\end{array} \right|}}{{\left| {({}^l{\bf{\bar p}}_{i - 1,m}^s - {}^l{\bf{\bar p}}_{i - 1,n}^s) \times ({}^l{\bf{\bar p}}_{i - 1,m}^s - {}^l{\bf{\bar p}}_{i - 1,o}^s)} \right|}} \eqno{(4)}
$$

\noindent where $ k $ represents the edge or surface feature indices in current frame. $ m,n,o $ represent the matching edge line or planar patch indices in last frame. For edge features $ {}^l{\bf{p}}_{i,k}^e $ in set $ {\bf{\psi }}_f^i $, $ {}^l{\bf{\bar p}}_{i - 1,m}^e $ and $ {}^l{\bf{\bar p}}_{i - 1,n}^e $ are the matching points in set $ {\bf{\bar \psi }}_c^{i - 1} $ that make up the edge line in last frame without motion-distortion. Similarly, $ {}^l{\bf{p}}_{i,k}^s $ is the surface feature in set $ {\bf{\zeta }}_f^i $. $ {}^l{\bf{\bar p}}_{i - 1,m}^s $, $ {}^l{\bf{\bar p}}_{i - 1,n}^s $, and $ {}^l{\bf{\bar p}}_{i - 1,o}^s $ are the points that consist of the corresponding planar patch in set $ {\bf{\bar \chi }}_c^{i - 1} $ or $ {\bf{\bar \zeta }}_c^{i - 1} $.

\subsection{Semantic lidar odometry}

Based on the perception results in Section 3.2, the odometry module firstly estimates the absolute increment for ego motion and relative motion increment of each semantic object in parallel. Then, each object pose is calculated in body-fixed coordinate \cite{henein2020dynamic}.

\subsubsection{Robust ego odometry}

Before entering the ego-motion estimation process, {\bf{Algorithm 1}} is used to achieve a geometric consistency check on the ground and background features, eliminating unknown dynamic influences that the object detection module cannot identify. The feature sets that passed the geometric consistency check are denoted as ($ {\bf{\tilde \chi }}_f^i $, $ {\bf{\tilde \psi }}_f^i $, $ {\bf{\tilde \zeta }}_f^i $).

\begin{algorithm}
        \caption{Geometric consistency feature check}
        \KwIn{Current feature sets: (${\bf{\chi }}_f^i$, ${\bf{\psi }}_f^i$, ${\bf{\zeta }}_f^i$), candidate sets without distortion: (${\bf{\bar \chi }}_c^{i - 1}$, ${\bf{\bar \psi }}_c^{i - 1}$, ${\bf{\bar \zeta }}_c^{i - 1}$), increment ${}_l^{i - 2}{{\bf{\delta }}^{i-1}}$ from last loop, association update interval $\lambda$, thresholds: ($t{h_{r}}$, $t{h_g}$, $t{h_b}$)}
        \KwOut{ Consistent feature sets: (${\bf{\tilde \chi }}_f^i$, ${\bf{\tilde \psi }}_f^i$, ${\bf{\tilde \zeta }}_f^i$)}
         
        The number of feature sets (${\bf{\chi }}_f^i$, ${\bf{\psi }}_f^i$, ${\bf{\zeta }}_f^i$) is recorded as ${N_f}$. Let ${}_l^{i - 2}{\bf{\delta }}_{best}^{i-1} = {}_l^{i - 2}{{\bf{\delta }}^{i-1}}$; \\
        \For{$i = 1:{N_{iter}}$}{
            \If{$i{\kern 1pt} {\kern 1pt} \% {\kern 1pt} {\kern 1pt} \lambda {\kern 1pt} {\kern 1pt}  = {\kern 1pt} {\kern 1pt} {\kern 1pt} 0$}
            {
                  Based on ${}_l^{i - 2}{\bf{\delta }}_{best}^{i-1}$ and (${\bf{\chi }}_f^i$, ${\bf{\psi }}_f^i$, ${\bf{\zeta }}_f^i$) including distortion correction time, feature position at the initial time of current frame is computed; \\
                  Search their closest points in kd-trees formed by (${\bf{\bar \chi }}_c^{i - 1}$, ${\bf{\bar \psi }}_c^{i - 1}$, ${\bf{\bar \zeta }}_c^{i - 1}$). Then correct (${\bf{\chi }}_f^i$, ${\bf{\psi }}_f^i$, ${\bf{\zeta }}_f^i$) to the end time of current frame, denoted as (${{\bf{m}}_{{{\bf{\chi }}_f}}}$, ${{\bf{m}}_{{{\bf{\psi }}_f}}}$, ${{\bf{m}}_{{{\bf{\zeta }}_f}}}$); \\
            }
        Random select matching pairs from ${{\bf{m}}_{{{\bf{\chi }}_f}}}$, ${{\bf{m}}_{{{\bf{\psi }}_f}}}$ and their candidate sets;\\
        Update ${}_l^{i - 2}{{\bf{\delta }}^{i-1}}$ with matching pairs and ICP model;\\
        Transform (${\bf{\chi }}_f^i$, ${\bf{\psi }}_f^i$, ${\bf{\zeta }}_f^i$) to the initial time of current frame with ${}_l^{i - 2}{{\bf{\delta }}^{i-1}}$. Find closest points in kd-trees to perform Euclidean distance check; \\
        Record the number of points passing the check as (${\eta _{{{\bf{\chi }}_f}}}$, ${\eta _{{{\bf{\psi }}_f}}}$, ${\eta _{{{\bf{\zeta }}_f}}}$). Let ${\eta _n} = {\eta _{{{\bf{\chi }}_f}}} + {\eta _{{{\bf{\psi }}_f}}} + {\eta _{{{\bf{\zeta }}_f}}}$; \\
            \If{$({\eta _n}/{N_f} \ge t{h_{r}}) \wedge ({\eta _{{{\bf{\chi }}_f}}} \ge t{h_g}) \wedge ({\eta _{{{\bf{\psi }}_f}}} + {\eta _{{{\bf{\zeta }}_f}}} \ge t{h_b}) \wedge ({\eta _n} > {\eta _{best}})$}
            {
                    Record feature sets (${\bf{\tilde \chi }}_f^i$, ${\bf{\tilde \psi }}_f^i$, ${\bf{\tilde \zeta }}_f^i$);\\
                    Let ${\eta _{best}} = {\eta _n}$, ${}_l^{i - 2}{\bf{\delta }}_{best}^{i-1} = {}_l^{i - 2}{{\bf{\delta }}^{i-1}}$;
            }
    }
\end{algorithm}

To improve the computational efficiency of ego odometry, we use the two-step algorithm proposed by \cite{shan2018lego} for pose estimation. First, frame increment $ [{t_z},{\theta _{roll}},{\theta _{pitch}}] $ is computed based on matched points in the checked ground feature set $ {\bf{\tilde \chi }}_f^i $, and the distortion-corrected candidate feature set $ {\bf{\bar \chi }}_c^{i - 1} $. Then, increment $ [{t_z},{\theta _{roll}},{\theta _{pitch}}] $ is fixed, and another 3-DOF increment $ [{t_x},{t_y},{\theta _{yaw}}] $ is estimated based on the edge feature $ {\bf{\tilde \psi }}_f^i $ , surface feature in $ {\bf{\tilde \zeta }}_f^i $ with more "flat" smoothness than $ {S_{a}} $ and their corresponding candidate sets. Finally, we obtain ego pose $ {\bf{T}}_{ol}^i $ by accumulating the 6-DOF increment $ {}_l^{i - 1}{{\bf{\delta }}^i} = [{t_x},{t_y},{t_z},{\theta _{roll}},{\theta _{pitch}},{\theta _{yaw}}] $.

\subsubsection{Semantic-geometric fusion multi-object tracking}

In order to track semantic objects robustly and accurately, a least-squares multi-object tracking module fusing semantic 3D-BB and geometric point clouds is designed in this paper. The module maintains a tracking list containing object information (object point cloud, sample points on 3D-BB, and poses in lidar coordinate) and tracking states (unique id, optimization quality evaluation and tracking continuity results) for object data association and relative motion increment estimation.

First, we use the constant motion model to predict object position, which, together with the object detection position, forms the position error term $ {{\bf{e}}_p} $. Meanwhile, the direction error term $ {{\bf{e}}_d} $ is formed by the relative object motion direction, and the 3D-BB size is used to calculate the scale error term $ {{\bf{e}}_s}$. Then the association matrix is solved by Kuhn-Munkres algorithm to obtain the object matching relationship. The calculation method of each error term can refer to \cite{bai2022apollo}.

For objects that are tracked in last frame but no matching relationship is currently found, we use the constant motion model to infer their positions in the subsequent two frames. If still no successful matching is found, they will be removed.

Then, we use the voxelized G-ICP algorithm proposed in \cite{koide2021voxelized} to construct the cost function of object point cloud $ {{\bf{\gamma }}^{i}} $ (object id $ k $ is omitted). Meanwhile, considering the point cloud distortion caused by the relative motion of object and robot between frames, we introduce a linear interpolation relative motion model to deal with this problem.

Assuming that in lidar coordinate $ l $, the $j$-th point on current processing object is $ {}^l{\bf{p}}_{i,j}^g $, and its corresponding timestamp is $ {t_j} $. Let $ {}_l^{i - 1}{\bf{H}}_j^i \in {\rm{SE}}(3) $ denote pose transformation between interval $\left[ {{t^{i - 1}},{t_j}} \right] $. Accordingly, $ {}_l^{i - 1}{\bf{H}}_j^i $ can be calculated by linear interpolation of the relative motion increment $ {}_l^{i - 1}{{\bf{H}}^i} \in {\rm{SE}}(3)$:

$$
{}_l^{i - 1}{\bf{H}}_j^i = \frac{{{t_j} - {t^i}}}{{{t^{i - 1}} - {t^i}}}{}_l^{i - 1}{{\bf{H}}^i} \eqno{(5)}
$$

We use ${}_l^{i - 1}{{\bf{R}}^i}$ and ${}_l^{i - 1}{{\bf{t}}^i}$ to represent the rotation and translation parts of the increment $ {}_l^{i - 1}{{\bf{H}}^i} $. The geometric error term $ {{\bf{e}}_{g_j}} $ containing motion distortion correction is expressed as:

$$
{{\bf{e}}_{g_j}} = \frac{{\sum {{}^l{\bf{p}}^g_{i - 1,m}} }}{{{N_j}}} - {}_l^{i - 1}{\bf{H}}_j^i{\kern 1pt} {\kern 1pt} {}^l{\bf{p}}_{i,j}^g \eqno{(6)}
$$

\noindent where $ {}^l{\bf{p}}_{i - 1,m}^g $ is the  $m$-th point in matched voxel $ {}^l{\bf{v}}_{i - 1}^g $ at the end time of the last frame, $ {N_j} $ is the number of 3D points contained in the matched voxel. The covariance matrix $ {{\bf{\Omega }}_j} $ corresponding to the error term $ {{\bf{e}}_{g_j}} $ is as follows:

$$
{{\bf{\Omega }}_j} = \frac{{\sum {C^g_{i - 1,m}} }}{{{N_j}}} - {}_l^{i - 1}{\bf{H}}_j^i{\kern 1pt} {\kern 1pt}C_{i,j}^g{\kern 1pt} {\kern 1pt}{}_l^{i - 1}{\bf{H}}{_j^{i{\kern 1pt} {\kern 1pt} }}^T \eqno{(7)}
$$

\noindent $ C_{i - 1,m}^g $  and $ C_{i,j}^g $ denote the covariance matrix. They describe the surrounding shape distribution of the point in the matched voxel and point $ {}^l{\bf{p}}_{i,j}^g $.

Note that objects detected in a single frame always have less constrained information. Meanwhile, the inaccurate initial value of the estimator may also lead to insufficient matchings at the beginning of the optimization. Eq. (6) sometimes fails to constrain the 6-DOF motion estimation. Therefore, while using geometric point cloud constraints, this paper further introduces semantic constraints to increment estimation for providing the right converged direction.

Specifically, we model a 3D-BB $ {{\bf{\tau }}^i} $ as a cuboid containing six planes and use $ {\bf{\pi }} = [{\bf{n}},d] $ to represent a single plane. Where, ${\bf{n}}$ is the plane normal vector with $ {\left\| {\bf{n}} \right\|_2} = 1 $. $ d $ is the distance from the origin to the plane. Unlike the geometric point cloud error with iterative matching for estimation, the matching relationship for 3D-BB plane features is directly available. Then, using the closest point $ {\bf{\Pi }} $ \cite{geneva2018lips} to parameterize $ {\bf{\pi }} $, i. e. $ {\bf{\Pi }} = {\bf{n}}d $, we can get the plane-to-plane semantic error term $ {{\bf{e}}_{{b_r}}} $ as follows:

$$
{{\bf{e}}_{{b_r}}} = {}^l{\bf{\Pi }}_{i - 1,r}^b - {}^l{\bf{\hat \Pi }}_{i - 1,r}^b \eqno{(8)}
$$

\noindent $ {}^l{\bf{\Pi }}_{i - 1,r}^b $ and $ {}^l{\bf{\hat \Pi }}_{i - 1,r}^b $ are the r-th measured and estimated planes on 3D-BB at time $ i $ in lidar coordinate $ l $. As described in \cite{hartley2003multiple}, the motion transformation of a plane can be expressed as:

$$
\left[ \begin{array}{l}
{}^l{\bf{n}}_{i - 1,r}^b\\
{}^ld_{i - 1,r}^b
\end{array} \right] = \left[ \begin{array}{l}
{\kern 1pt} {\kern 1pt} {\kern 1pt} {\kern 1pt} {\kern 1pt} {\kern 1pt} {\kern 1pt} {\kern 1pt} {\kern 1pt} {\kern 1pt} {\kern 1pt} {\kern 1pt} {\kern 1pt} {\kern 1pt} {\kern 1pt} {\kern 1pt} {\kern 1pt} {\kern 1pt} {\kern 1pt} {\kern 1pt} {\kern 1pt} {\kern 1pt} {\kern 1pt} {\kern 1pt} {}_l^{i - 1}{{\bf{R}}^i}{\kern 1pt} {\kern 1pt} {\kern 1pt} {\kern 1pt} {\kern 1pt} {\kern 1pt} {\kern 1pt} {\kern 1pt} {\kern 1pt} {\kern 1pt} {\kern 1pt} {\kern 1pt} {\kern 1pt} {\kern 1pt} {\kern 1pt} {\kern 1pt} {\kern 1pt} {\kern 1pt} {\kern 1pt} {\kern 1pt} {\kern 1pt} {\kern 1pt} {\kern 1pt} {\kern 1pt} {\kern 1pt} {\kern 1pt} {\kern 1pt} {\kern 1pt} {\kern 1pt} {\kern 1pt} {\kern 1pt} {\kern 1pt} {\kern 1pt} {\kern 1pt} {\bf{0}}\\
 - {({}_l^{i - 1}{{\bf{R}}^{i{\kern 1pt} {\kern 1pt} T}}{\kern 1pt} {\kern 1pt} {}_l^{i - 1}{{\bf{t}}^i})^T}{\kern 1pt} {\kern 1pt} {\kern 1pt} {\kern 1pt} {\kern 1pt} {\kern 1pt} {\kern 1pt} {\kern 1pt} {\kern 1pt} 1
\end{array} \right]\left[ \begin{array}{l}
{}^l{\bf{n}}_{i,r}^b\\
{}^ld_{i,r}^b
\end{array} \right] \eqno{(9)}
$$

Taking Eq.(9) into $ {}^l{\bf{\hat \Pi }}_{i - 1,r}^b $, we can further obtain:

$$
{}^l{\bf{\hat \Pi }}_{i - 1,r}^b = \left( {{}_l^{i - 1}{\bf{R}}_j^i{}^l{\bf{n}}_{i,r}^b{\kern 1pt} {\kern 1pt} } \right)\left( { - {}_l^{i - 1}{\bf{t}}{{_j^i}^T}{}_l^{i - 1}{\bf{R}}_j^i{}^l{\bf{n}}_{i,r}^b + d} \right) \eqno{(10)}
$$

When using the least squares method to estimate $ {}_l^{i - 1}{{\bf{H}}^i} $, it is necessary to obtain the derivative of $ {{\bf{e}}_{{b_r}}} $ with respect to $ {}_l^{i - 1}{{\bf{H}}^i} $:

$$
\frac{{\partial {{\bf{e}}_{{b_r}}}}}{{\partial {\delta _{{}_l^{i - 1}{\bf{\theta }}_j^i}}}}= -{\left[ {{\bf{F}}{}_l^{i - 1}{\bf{t}}{{_j^i}^T}{\bf{F}}} \right]_ \times }- {\bf{F}{}_l^{i - 1}{\bf{t}}{{_j^i}^T } {\left[ {\bf{F}} \right]_ \times } }
+ {\left[ {{\bf{F}}d} \right]_ \times } \eqno{(11)}
$$

$$
\frac{{\partial {{\bf{e}}_{{b_r}}}}}{{\partial {\delta _{{}_l^{i - 1}{\bf{t}}_j^i}}}} = {\bf{F}}{{\bf{F}}^T} \eqno{(12)}
$$

\noindent where ${\bf{F}} = {}_l^{i - 1}{\bf{R}}_j^i{}^l{\bf{n}}_{i,r}^b $. $ \frac{{\partial {{\bf{e}}_{{b_r}}}}}{{\partial {\delta _{{}_l^{i - 1}{\bf{\theta }}_j^i}}}} $ and $ \frac{{\partial {{\bf{e}}_{{b_r}}}}}{{\partial {\delta _{{}_l^{i - 1}{\bf{t}}_j^i}}}} $ represent the jacobians of the error term $ {{\bf{e}}_{{b_r}}} $ w.r.t. the rotation and translation component. The detailed derivation of Eq. (11) and (12) can be found in APPENDIX. 

Finally, Eq. (13) is the comprehensive cost function for the estimation of $ {}_l^{i - 1}{{\bf{H}}^i} $ considering the semantic confidence $\mu ^i$:

$$
\mathop {\arg \min }\limits_{{}_l^{i - 1}{{\bf{H}}^i}} \frac{{2 - {\mu ^i}}}{{2{N_g}}}\sum\limits_{j = 1}^{{N_g}} {\left( {{N_j} {\kern 1pt} {\kern 1pt}{\bf{e}}_{g_j}^T{\bf{\Omega }}_j^{ - 1}{{\bf{e}}_{g_j}}} \right)}  + \frac{{{\mu ^i}}}{{2{N_b}}}\sum\limits_{r = 1}^{{N_b}} {\left( {{\bf{e}}_{b_r}^T{{\bf{e}}_{b_r}}} \right)}  \eqno{(13)}
$$

\noindent where $ {N_g} $ is the number of object point cloud.  $ {N_b} $ is the number of plane features on 3D-BB.

In this paper, we use LM \cite{madsen2004methods} algorithm to solve Eq. (9) and the estimated relative increment to get motion-distorted object point cloud $ {{\bf{\bar \gamma }}^i} $ and object edge feature $ {\bf{\bar \gamma }}_c^i $, which will be used for object tracking in next frame and accumulated error elimination in mapping module.

\begin{figure*}[!htb]
\centering
\includegraphics[scale=0.17]{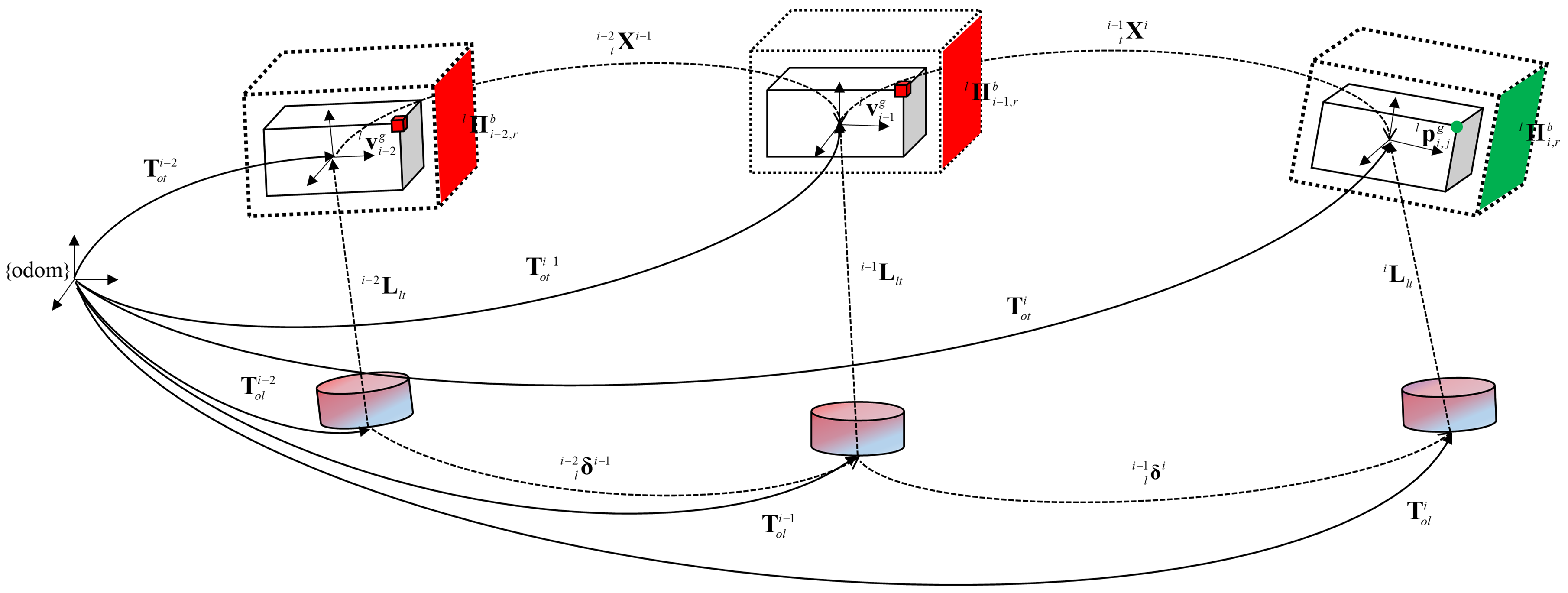}
\caption{ {\bf{Notation and coordinate frames.}} Coloured cylinders denote the lidar. White cuboids represent a moving object. Dashed cuboids are the result inferred by semantic detection module. Solid lines represent lidar and object pose in odom coordinate. Dashed lines indicate their motion increment in local coordinate. Small red cubes and parallelograms represent voxels whose motion distortion has been eliminated and 3D planes obtained by 3D-BB. Green dot and plane represent object points in current point cloud and 3D-BB plane feature without motion distortion.}
\label{fig_framework3}
\end{figure*}

\subsubsection{Object odometry}

Based on the calculation results of ego odometry and multi-object tracking, object pose in odom coordinate can be computed. First, object motion increment $ {}_l^i{{\bf{X}}^{i - 1}} $ in lidar coordinate is denoted as:

$$
{}_l^i{{\bf{X}}^{i - 1}} = {}_l^{i - 1}{{\bf{\delta }}^i}{\kern 1pt} {\kern 1pt}{}_l^i{{\bf{H}}^{i - 1}} \eqno{(14)}
$$

Then, combined with the body-fixed coordinate proposed in \cite{henein2020dynamic} (as shown in Fig.4), and assuming that the transform matrix from object to lidar at time $ i-1 $ is $ {}^{i - 1}{{\bf{L}}_{lt}} $, we can obtain absolute object motion increment ${}_t^{i - 1}{{\bf{X}}^i} $ in body-fixed coordinate:

$$
{}_t^{i - 1}{{\bf{X}}^i} = {}^{i - 1}{\bf{L}}_{lt}^{ - 1}{\kern 1pt} {\kern 1pt}{}_l^i{{\bf{X}}^{i - 1}}{\kern 1pt} {\kern 1pt}{}^{i - 1}{{\bf{L}}_{lt}} \eqno{(15)}
$$

Odometry pose ${\bf{T}}_{ot}^i $ for each object at time $i$ can be computed by accumulating increment $ {}_t^{i - 1}{{\bf{X}}^i} $. This paper uses ego odometry $ {{\bf{T}}_{ol}} $ and transform matrix $ {{\bf{L}}_{lt}} $ extracted from semantic detection module to set object odometry $ {{\bf{T}}_{ot}} $ at the first detected frame.

\subsection{4D scene mapping}

To reuse the created maps, we suggest maintaining the long-term static map and movable object maps separately. First, we use the object absolute trajectory tracking list (ATTL) to detect dynamic objects. Static map creation aims to update the correction matrix $ {\bf{T}}_{mo}^i $ for accumulated error correction with static objects and environment features. Finally, we correct the object pose in map coordinate and update the object map based on the object motion state.

\subsubsection{Dynamic object detection}

As shown in Fig.5, the ATTL maintained by MLO system in the mapping module is used to analyze the motion state of each instance semantic object. First, the current frame objects are matched with the ATTL through the unique id obtained from SGF-MOT module. Then, we use the current error correction matrix $ {\bf{T}}_{mo}^{i - 1} $ to predict the object pose $ {\bf{T}}_{mt}^{i\;\;'} $ in map coordinate:

$$
{\bf{T}}_{mt}^{i\;\;'} = {\bf{T}}_{mo}^{i - 1}{\bf{T}}_{ot}^i \eqno{(16)}
$$

By calculating the motion increment of the predicted object pose and the oldest static pose of the matching object in ATTL, we can determine whether each object is in motion. Note that the detected object in the first frame is set to be static by default.

\begin{figure}[t]
\centering
\includegraphics[scale=0.22]{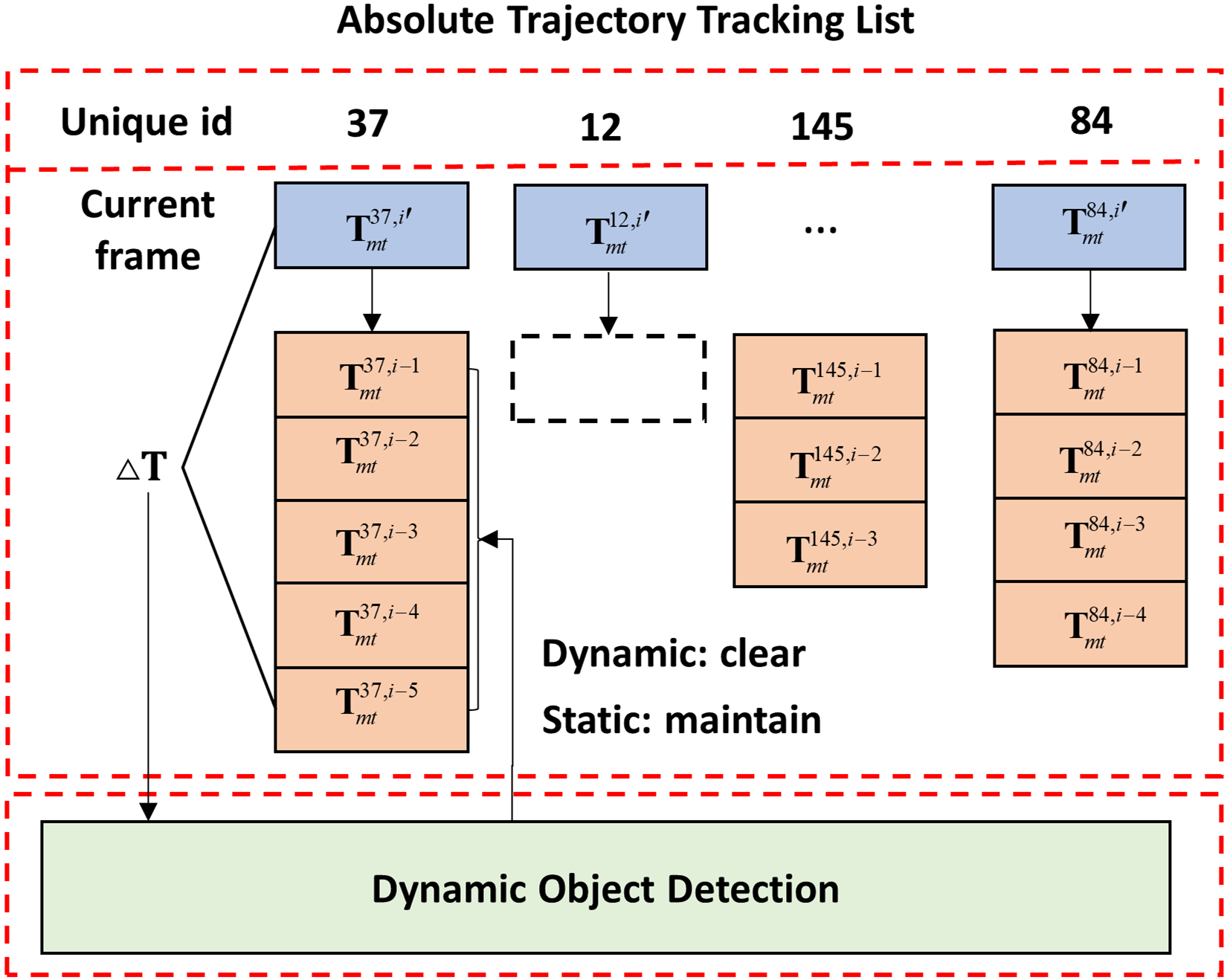}
\caption{{\bf{Dynamic object detection based on ATTL.}} The blue boxes represent the object predicted pose in the current frame, and the orange boxes represent the object tracking list that has corrected the accumulated error. Objects with ids 37 and 84 are in normal tracking state. The object with id 12 will be initialized as a static object, and the object with id 145 will be deleted because the current frame does not observe it. $ \Delta {\bf{T}} $ is used to determine the current motion state of each object. }
\label{fig_framework}
\end{figure}

\subsubsection{Static map creation} 

By matching the static object features $ {\bf{\bar \gamma }}_c^i $ in the current frame with the temporary object map, ground feature ${\bf{\bar \chi }}_c^i $, background edge feature $ {\bf{\bar \psi }}_c^i $ and surface feature $ {\bf{\bar \zeta }}_c^i $ with their surrounding point cloud map, ego pose $ {\bf{T}}_{ol}^i $ can be corrected to map coordinate. Then, a long-term static map can be built by aligning the non-semantic features into map coordinate. Readers can refer to \cite{zhang2014loam} for more map matching and optimization details.

According to the ego pose $ {\bf{T}}_{ol}^i $ in odom coordinate and the corrected pose $ {\bf{T}}_{ml}^i $ in map coordinate, the updated accumulated error correction matrix $ {\bf{T}}_{mo}^i $ can be obtained.

\subsubsection{Tracking list maintain}

The ATTL can be updated based on the motion state of each object. If the object is moving, the map should be cleared first. Otherwise, the map will remain unchanged. Then, we use $ {\bf{T}}_{mo}^i $ and $ {\bf{T}}_{ml}^i $ to convert the object odometry pose $ {\bf{T}}_{ot}^i $ and extracted object features $ {\bf{\bar \gamma }}_c^{k,i} $ into map coordinate, which will participate in subsequent mapping steps.

\begin{figure}[h]
\centering
\includegraphics[scale=0.215]{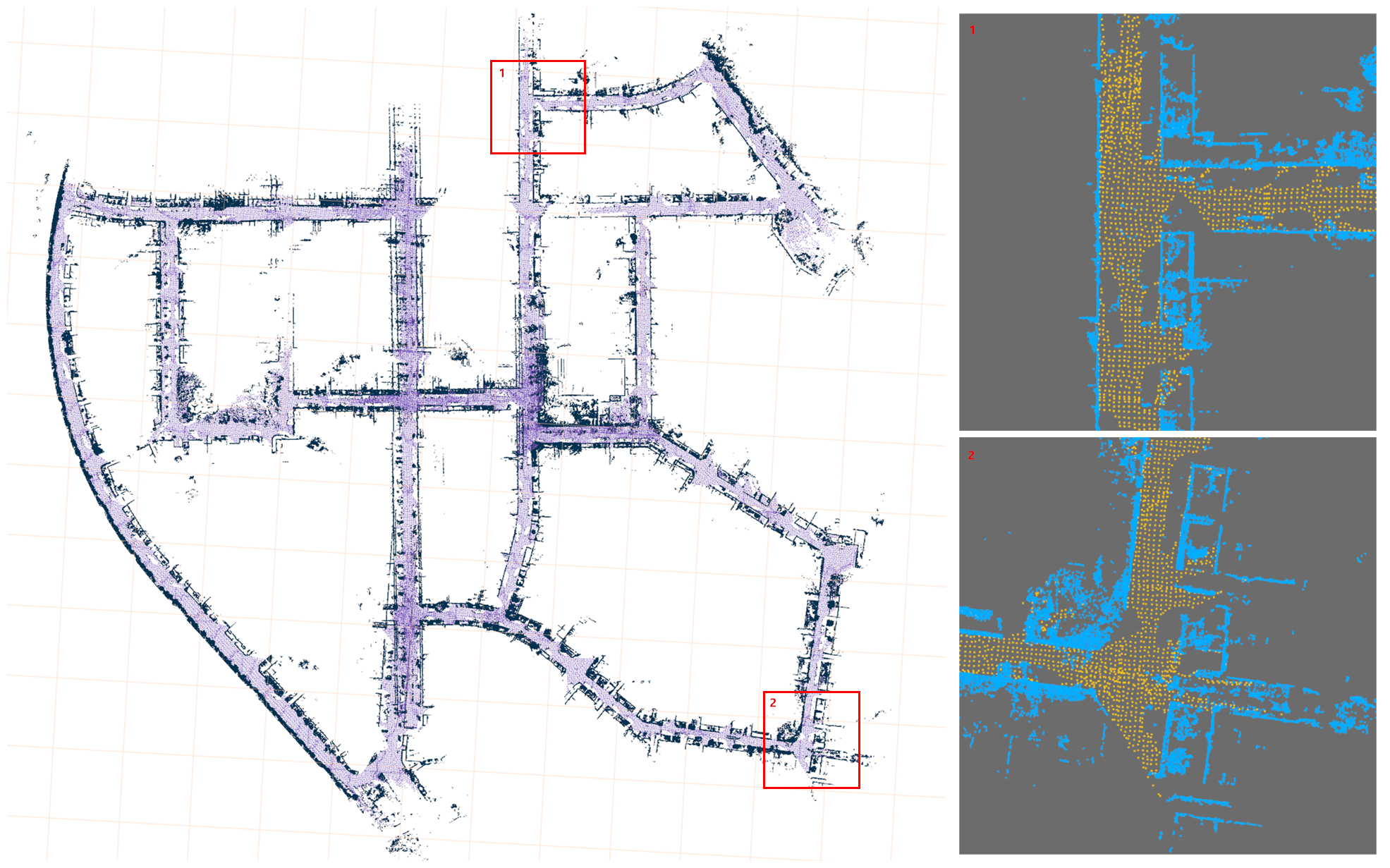}
\caption{{\bf{Point cloud map created by sequence 00 in the KITTI-odometry dataset.}}  Blue point cloud on the left represents the background map, and purple point cloud represents the ground. As shown in the enlarged section, the resulting map does not contain semantic objects that may change over time by maintaining semantic objects and static map separately.}
\label{fig_framework}
\end{figure}

\section{Experiment}

The experimental environment of the proposed framework is AMD® Ryzen 7 5800h (8 cores @3.2 GHz), 16GB RAM, ROS Melodic, and NVIDIA GeForce RTX 3070. We use absolute trajectory error (ATE) \cite{sturm2012benchmark}, relative rotation error (RRE) and relative translation error (RTE) \cite{geiger2012we} to evaluate the accuracy and odometry drift of ego and object localization. The CLEAR MOT metric \cite{bernardin2008evaluating} is used to evaluate estimated accuracy and tracking object configurations consistently over time for multi-object tracking module.

Note that our multi-object tracking module is performed in lidar coordinate. Therefore, the relative motion tracking ability of the object is evaluated. On the other hand, object trajectories in map coordinate evaluate the overall accuracy of ego localization and multi-object tracking module.

Furthermore, the point cloud in the KITTI-raw benchmark does not use known GPS/IMU information to correct motion distortion nor provides ground-truth trajectories that can be directly used for localization evaluation. For obtaining a smooth and accurate 6-DOF pose ground truth, we use the extended Kalman filter \cite{moore2016generalized} to process GPS/IMU data in the KITTI raw dataset. Combined with the high-precision artificial object annotation in lidar coordinate, the ground truth of object pose in map coordinate can be obtained.

\subsection{Ego odometry localization accuracy experiment}

The KITTI-odometry benchmark contains rich static objects with long sequences. Meanwhile, the KITTI-raw dataset contains more challenging scenes, such as the city sequence with unknown dynamic semantic objects (trains). We evaluate the MLO system in these two datasets by comparing it with other state-of-the-art 3D lidar SLAM. Since the KITTI dataset labels semantic objects within the camera's field of view, only the ROI-cutting lidar data is input into the fusion perception module.

\begin{table}[h]\scriptsize
\caption{ Ego localization accuracy under KITTI-odometry benchmark. }
\centering
\begin{tabular}{c c c c c c}
\toprule  
\multicolumn{6}{c}{\bf{Absolute Motion Trajectory RMSE [m]}}\\
\toprule  
\bf{Seq} & \bf{A-LOAM} & \bf{Lego-LOAM} & \bf{F-LOAM} & \bf{MLO-ddo} & \bf{MLO} \\
\specialrule{0em}{0pt}{2pt}
\toprule  
00 & 12.12 & 41.99 & 14.71 & 8.23 & \bf{7.52} \\
01 & 15.81 & {\bf{fail}} & 17.61 & \bf{12.11} & 12.42 \\
02 & {\bf{fail}} & {\bf{fail}} & {\bf{fail}} & 18.38 & \bf{17.65} \\
03 & 0.57 & 14.19 & 0.90 & 0.52 & \bf{0.50} \\
04 & 0.45 & {\bf{fail}} & 0.36 & 0.29 & \bf{0.27} \\
05 & 5.72 & 8.12 & 8.06 & 2.74 & \bf{1.92} \\
06 & \bf{0.44} & {\bf{fail}} & 1.04 & 0.45 & 0.46 \\
07 & 2.73 & 3.75 & 2.49 & 2.33 & \bf{1.48} \\
08 & \bf{4.42} & {\bf{fail}} & 6.02 & 5.18 & 4.86 \\
09 & \bf{3.48} & {\bf{fail}} & 21.53 & 4.66 & 4.72 \\
10 & \bf{1.18} & 10.57 & 1.54 & 1.80 & 1.90 \\
\hline
\specialrule{0em}{0pt}{2pt}
Mean & 4.69 & {\bf{fail}} & 7.43 & 3.88 & \bf{3.66} \\
\hline
\end{tabular}
\end{table}

For the experimental lidar SLAM system, A-LOAM \cite{qin2019aloam} is a simplified version of LOAM \cite{zhang2014loam} with the removal of the IMU used. It also extracts edge and surface features from point cloud to perform frame-to-frame matching for odometry estimation and frame-to-map matching to complete mapping. Lego-LOAM \cite{shan2018lego} takes into account ground constraints. Use ground features and background edge features, respectively, to estimate the increment $ [{t_z},{\theta _{roll}},{\theta _{pitch}}] $ and $ [{t_x},{t_y},{\theta _{yaw}}] $ for improving efficiency. F-LOAM \cite{wang2021f} proposes a two-step motion distortion removal algorithm, and then localization and mapping are done directly through frame-to-map matching. 

As shown in Table I, our method successfully runs all sequences under the KITTI-odometry benchmark and achieves the best results in most cases. When estimating $ [{t_x},{t_y},{\theta _{yaw}}] $ based on background edge features, the Lego-LOAM \cite{shan2018lego} system shows severe drift on many sequences due to only using point clouds from the camera's view. Our system also uses the two-step estimation algorithm to ensure computational efficiency. However, using background surface features based on smoothness prior $ {S_a} $ and geometric consistency check algorithm improve the MLO system robustness. 

\begin{table*}[h] \normalsize
\caption{Evaluation for ego localization accuracy under KITTI-raw City and Residential sequences.}
		\newcolumntype{C}{>{\centering\arraybackslash}X}
		\begin{tabular}{c ccc ccc ccc}
		\toprule  
		\multirow{2}*{\bf{Sequence}} & \multicolumn{3}{c }{\bf{A-LOAM}}& \multicolumn{3}{c }{\bf{MLO-gc}}& \multicolumn{3}{c}{\bf{MLO}}\\
		& ATE[m] & RTE[m/f] & RRE[deg/f] & ATE[m] & RTE[m/f] & RRE[deg/f] & ATE[m] & RTE[m/f] & RRE[deg/f] \\
		\specialrule{0em}{0pt}{2pt}
		\toprule  
		0926-0011 & \bf{0.36} & 0.09 & 0.11 & 0.39 & 0.06 & 0.08 & 0.39 & 0.06 & 0.07 \\
		0926-0014 & 0.42 & 0.11 & 0.21 & \bf{0.26} & 0.11 & 0.14 & 0.27 & 0.11 & 0.13 \\
		0926-0056 & 0.37 & 0.16 & 0.21 & {\bf{fail}} & 0.51 & 0.15 & \bf{0.32} & 0.12 & 0.13 \\
		0926-0059 & \bf{0.24} & 0.06 & 0.14 & 0.28 & 0.07 & 0.11 & 0.28 & 0.06 & 0.10 \\
		0929-0071 & \bf{1.62} & 0.05 & 0.15 & 1.63 & 0.07 & 0.14 & 1.63 & 0.06 & 0.12 \\
		0926-0019 & 5.56 & 0.32 & 0.16 & \bf{2.46} & 0.25 & 0.13 & 2.57 & 0.25 & 0.13 \\
		0926-0022 & 1.83 & 0.08 & 0.21 & \bf{1.75} & 0.11 & 0.20 & 1.81 & 0.08 & 0.16 \\
		0926-0039 & 0.53 & 0.07 & 0.28 & 0.43 & 0.12 & 0.23 & \bf{0.43} & 0.06 & 0.18 \\
		0926-0064 & 1.76 & 0.07 & 0.24 & 1.62 & 0.07 & 0.18 & \bf{1.60} & 0.07 & 0.18 \\
        \hline
		\specialrule{0em}{0pt}{2pt}
		Mean & 1.41 & 0.11 & 0.19 & 2.99 & 0.15 & 0.15 & \bf{1.03} & \bf{0.10} & \bf{0.13} \\
        \hline
		\end{tabular}
\end{table*}

Further, detecting dynamic objects by ATTL and adding static object constraints for static map creation can better eliminate accumulated error compared to MLO-ddo system, which does not detect and use static semantic objects in the mapping module. In sequences 00, 02, 03, 05  and 07 containing rich static objects, the MLO system achieved better localization accuracy.

Table II shows that in complex scenes, the MLO system can still achieve better localization accuracy in most cases. Meanwhile, the MLO system outperforms the other two frameworks in terms of average RRE and RTE. For sequences 0926-0056, there is a train that the object detection module cannot recognize. Since the A-LOAM \cite{qin2019aloam} system does not distinguish the ground and background features and makes use of all the information for localization and mapping, it can improve the system robustness to a certain extent. For the MLO system that also uses the two-step estimation algorithm, especially for the estimation of $ [{t_x},{t_y},{\theta _{yaw}}] $, a geometric consistency check can reduce the proportion of abnormal associations and avoid the trajectory drift problem of the MLO-gc system (ego-motion estimation module uses the features extracted after removing the object point cloud, but does not consider the geometric consistency check algorithm).

\begin{figure}[h]
\centering
  \subfigure[Sequence 0001]{
    \label{fig:subfig:a} 
    \includegraphics[scale=0.22]{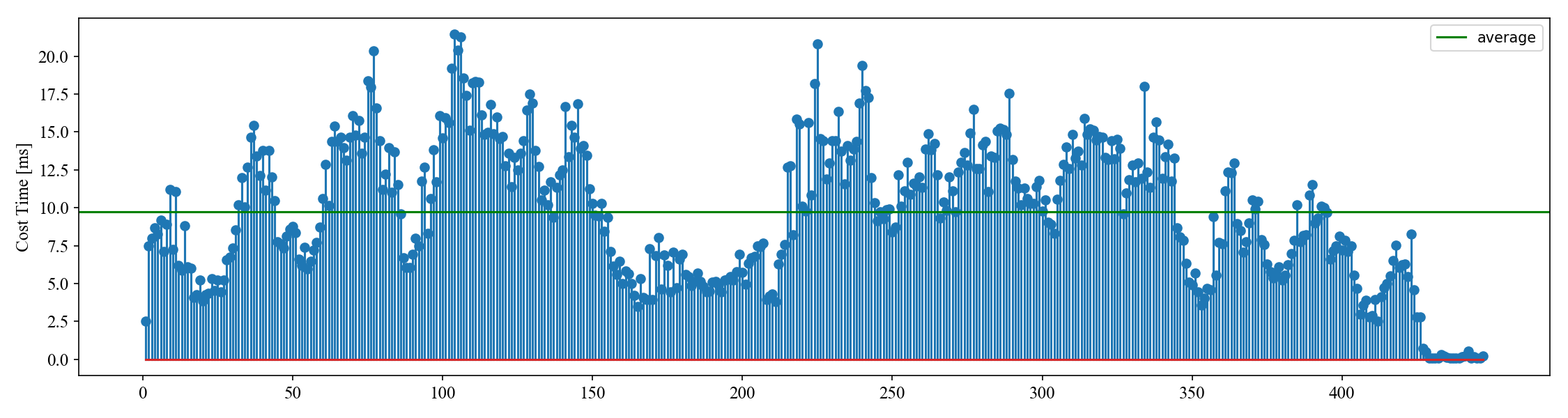}}
  \subfigure[Sequence 0015]{
    \label{fig:subfig:c} 
    \includegraphics[scale=0.22]{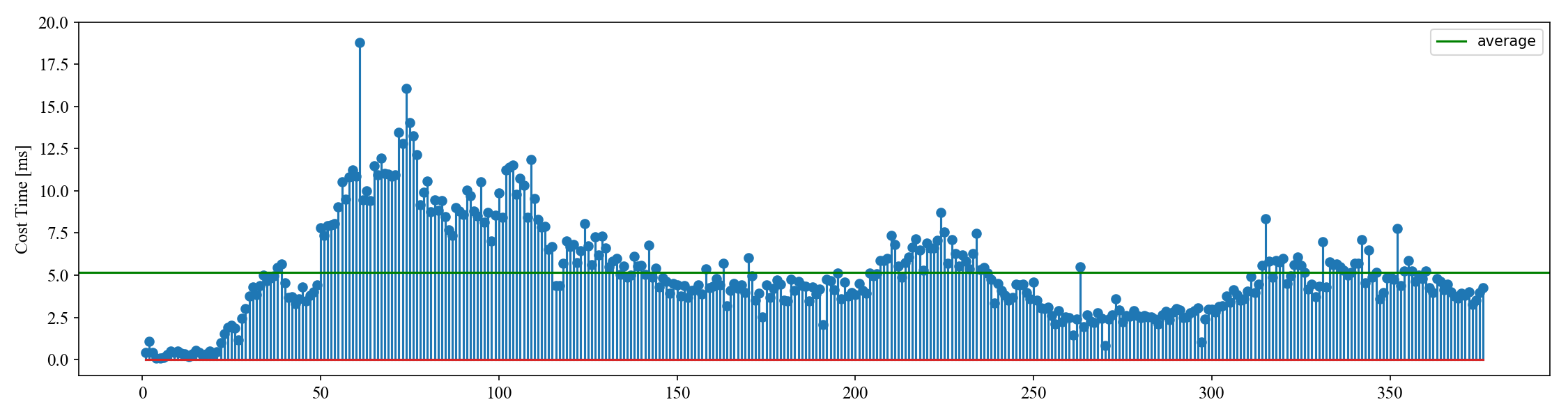}}
\caption{Time-consuming results of the semantic-geometric fusion tracking module in sequences filled with objects under the KITTI-tracking dataset.}
\label{fig_framework5}
\end{figure}

\subsection{Object tracking accuracy and robustness experiment}

First, we compare the SGF-MOT module proposed in this paper with AB3DMOT \cite{weng20203d} and PC3T \cite{wu20213d} trackers under the KITTI-tracking benchmark. The tracking results are evaluated in 2D image, which can be computed by the KITTI calibration matrix. AB3DMOT \cite{weng20203d} only takes point cloud as input and uses the Kalman filter for 3D-BB tracking in lidar coordinate. PC3T \cite{wu20213d} uses the ground truth of ego localization provided by KITTI dataset to transform the detected object into map coordinate. Then, it performs absolute motion tracking based on the object kinetic model.

\begin{table}[h] \small
\caption{ Evaluation for multi-object tracking module under the KITTI-tracking benchmark. }
\centering
\begin{tabular}{c c c c}
\toprule  
{ } & $ {\bf{MOT}}{{\bf{A}}_{2d}} $ & $ {\bf{MOT}}{{\bf{P}}_{2d}} $ & {\bf{Time}} [ms] \\
\specialrule{0em}{0pt}{2pt}
\bottomrule 
\specialrule{0em}{0pt}{2pt}
\bf{AB3DMOT} & 68.83 & 87.23  & 4.82 \\
\bf{PC3T}  & \bf{85.33}   & 87.18 & \bf{4.5} \\
\specialrule{0em}{0pt}{2pt}
\hline
\specialrule{0em}{0pt}{2pt}
\bf{SGF-MOT}  & 77.55  & \bf{87.27} & 5.01 \\
\hline
\end{tabular}
\end{table}

As shown in Table III, thanks to the accurate point cloud, three methods have little difference in the MOTP metric, which illustrates the estimation accuracy of the tracker. Since the object is transformed into map coordinate using the prior ego ground truth, PC3T \cite{wu20213d} tracker only needs to consider the uncertainty caused by absolute object motion between frames. Moreover, it achieves the best results on the MOTA metric. The proposed SGF-MOT tracker does not use ego localization information but also improves the tracking robustness by introducing a least-squares estimator fusing geometric and semantic constraints. So we achieve better MOTA results than the AB3DMOT \cite{weng20203d} tracker. 

\begin{figure}[t]
\centering
  \subfigure[Semantic only MOT method]{
    \label{fig:subfig:a} 
    \includegraphics[scale=0.16]{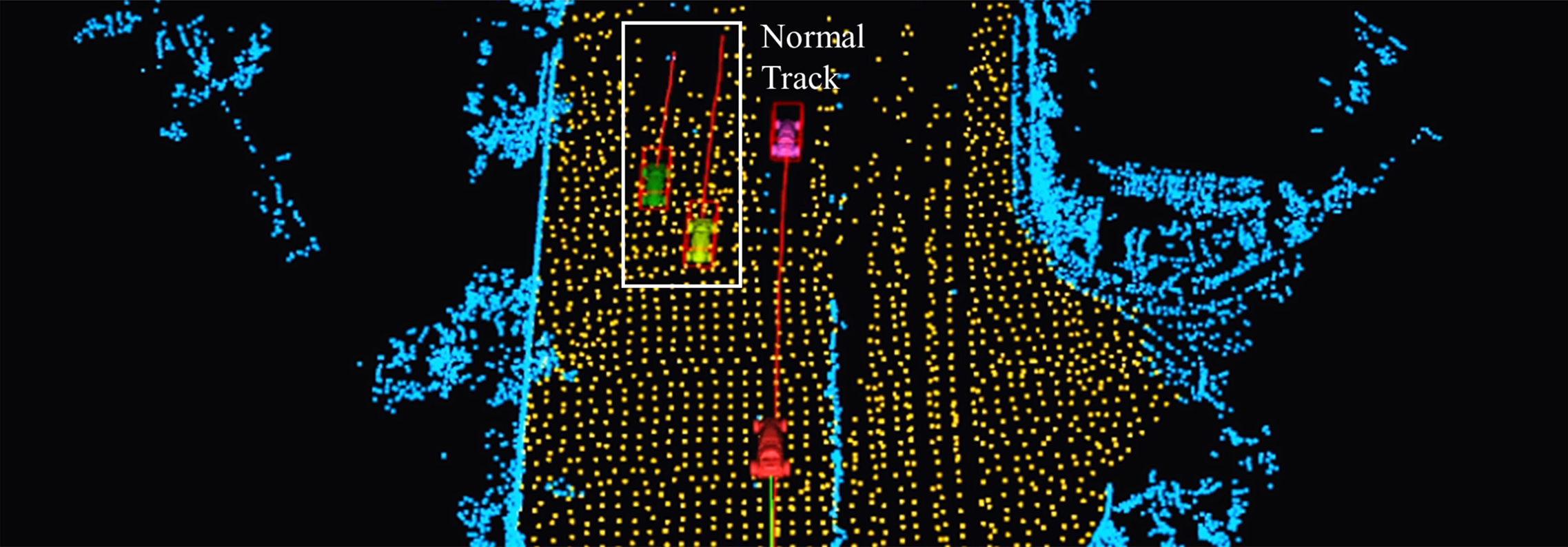}}
  \subfigure[Geometry only MOT method]{
    \label{fig:subfig:b} 
    \includegraphics[scale=0.16]{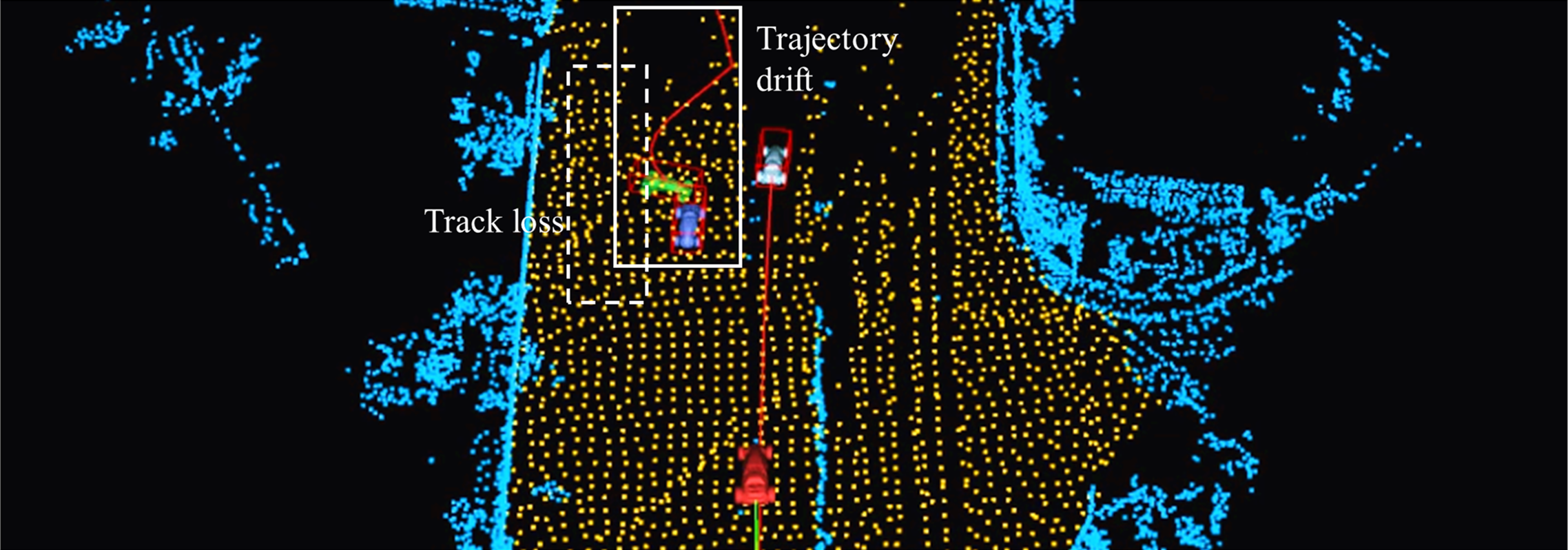}}
  \subfigure[Fusion MOT Method]{
    \label{fig:subfig:c} 
    \includegraphics[scale=0.16]{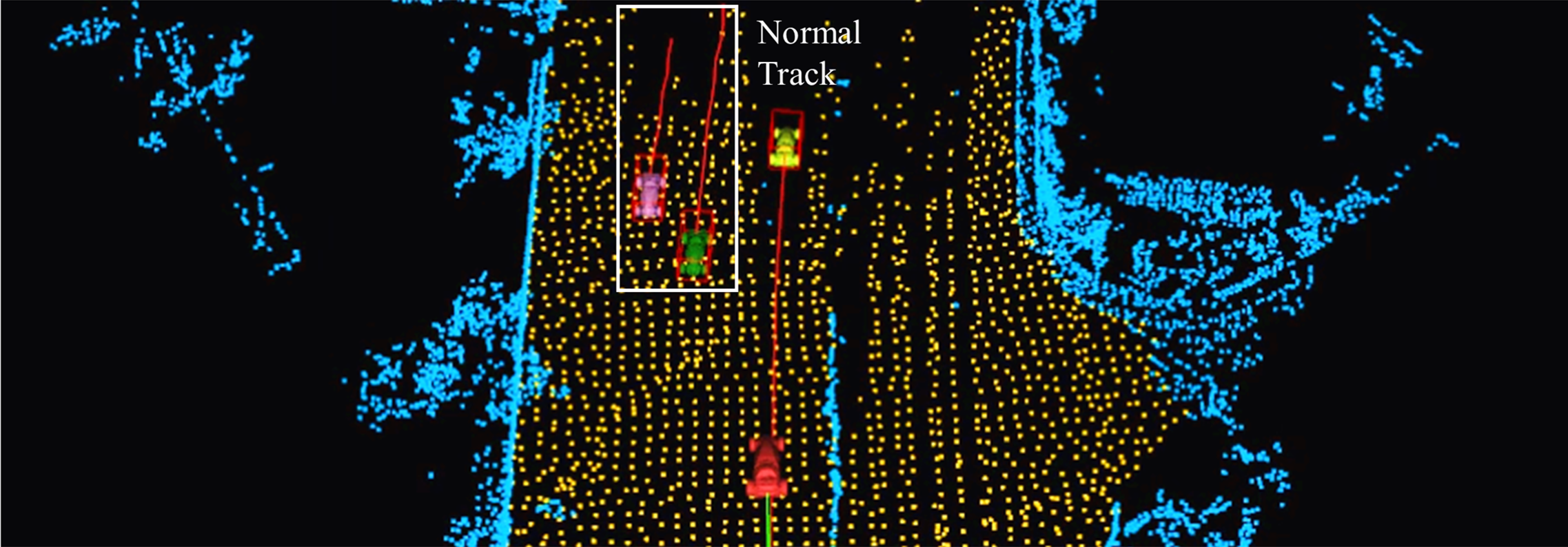}}
\caption{Comparison of different tracking methods in sequence 0926-0056 under the KITTI-tracking dataset. Both semantic only and fusion methods can track objects robustly, but geometric only method have the problems of object tracking loss and trajectory drift.}
\label{fig_framework7}
\end{figure}

Like the other two methods, our method runs in real-time on the CPU. Fig.7 shows the time-consuming results when the fusion tracking module executes sequence 0001 and 0015. It can be concluded that the time-consuming of our method is always less than 25ms, even during peak hours.

Then, we selected 8 sequences containing moving objects from KITTI-raw City and Road sequences. The semantic-geometric fusion tracking method (Fusion Method) is compared with methods using only semantic boxes (Semantic Only) and geometric point clouds (Geometry Only) by evaluating the object trajectory accuracy in map coordinate. 

\begin{table*}[h] \small
\caption{Evaluation for absolute object trajectory accuracy under KITTI-raw City and Road sequences.}
		\newcolumntype{C}{>{\centering\arraybackslash}X}
		\centering
		\begin{tabular}{c c ccc ccc ccc}
		\toprule  
		\multirow{2}*{\bf{Sequence}} & \multirow{2}*{\bf{id}} & \multicolumn{3}{c }{\bf{Geometry Only}}& \multicolumn{3}{c }{\bf{Semantic Only}}& \multicolumn{3}{c}{\bf{Fusion Method}}\\
		& {} & ATE[m] & RTE[m/m]& RRE[deg/m] & ATE[m] & RTE[m/m] & RRE[deg/m] & ATE[m] & RTE[m/m] & RRE[deg/m] \\
		\specialrule{0em}{0pt}{2pt}
		\toprule  
		\specialrule{0em}{0pt}{2pt}
		\multirow{2}*{0926-0009} & 87 & \bf{0.11} & 0.11 & 0.35 & 0.25 & 0.33 & 2.10 & 0.17 & 0.13 & 0.36 \\
		 & 89 & \bf{0.06} & 0.06 & 0.32 & 0.16 & 0.16 & 1.55 & 0.14 & 0.11 & 0.86 \\
		\specialrule{0em}{0pt}{2pt}
		\cline{1-2}
		\specialrule{0em}{0pt}{2pt}
		\multirow{2}*{0926-0013} & 1 & \bf{0.21} & 0.15 & 0.93 & 0.39 & 0.29 & 3.85 & 0.40 & 0.26 & 2.76 \\
		 & 2 & {\bf{\color{red}2.54}} & 1.03 & 9.80 & \bf{0.22} & 0.18 & 1.50 & 0.23 & 0.16 & 1.30 \\
		\specialrule{0em}{0pt}{2pt}
		\cline{1-2}
		\specialrule{0em}{0pt}{2pt}
		\multirow{2}*{0926-0018} & 6 & 0.11 & 0.12 & 1.77 & 0.12 & 0.13 & 2.53 & \bf{0.08} & 0.12 & 1.09 \\
		 & 11 & \bf{0.08} & 0.10 & 1.39 & 0.15 & 0.16 & 1.90 & \bf{0.11} & 0.12 & 1.02 \\
		\specialrule{0em}{0pt}{2pt}
		\cline{1-2}
		\specialrule{0em}{0pt}{2pt}
		\multirow{2}*{0926-0051} & 21 & {\bf{\color{red}2.28}} & 1.86 & 2.83 & 0.20 & 0.18 & 1.73 & \bf{0.18} & 0.13 & 1.22 \\
		 & 26 & \bf{0.07} & 0.12 & 0.27 & 0.14 & 1.54 & 1.21 & 0.09 & 0.09 & 1.13 \\
		\specialrule{0em}{0pt}{2pt}
		\cline{1-2}
		\specialrule{0em}{0pt}{2pt}
		0926-0084 & 14 & {\bf{\color{red}4.81}} & 1.11 & 7.11 & 0.27 & 0.38 & 6.25 & \bf{0.23} & 0.35 & 6.14 \\
		\specialrule{0em}{0pt}{1pt}
		\cline{1-2}
		\specialrule{0em}{0pt}{2pt}
		\multirow{2}*{0926-0015} & 32 & \bf{0.14} & 0.25 & 1.67 & 0.19 & 0.21 & 2.56 & 0.16 & 0.16 & 1.96 \\
		 & 35 & {\bf{\color{red}3.29}} & 1.07 & 2.28 & 0.30 & 0.28 & 5.44 & \bf{0.27} & 0.27 & 7.15 \\
		\specialrule{0em}{0pt}{1pt}
		\cline{1-2}
		\specialrule{0em}{0pt}{2pt}
		0926-0028 & 1 & {\bf{fail}} & {\bf{fail}} & {\bf{fail}} & 0.31 & 0.22 & 2.22 & \bf{0.31} & 0.21 & 2.06 \\
		\specialrule{0em}{0pt}{2pt}
		\cline{1-2}
		\specialrule{0em}{0pt}{2pt}
		0926-0032 & 15 & 0.18 & 0.16 & 0.43 & 0.20 & 0.21 & 1.06 & \bf{0.17} & 0.18 & 1.31 \\
        \hline
		\specialrule{0em}{0pt}{2pt}
		\multicolumn{2}{c }{Mean} & 1.16 & 0.52 & 2.43 & 0.22 & 0.34 & 2.64 & \bf{0.19} & \bf{0.17} & \bf{2.19} \\
		\bottomrule 
		\end{tabular}
\end{table*}

The object localization accuracy of geometry-only method exhibit a distinct "bipolar" distribution. Furthermore, we found that it can only achieve stable and accurate tracking for objects perceived within 15m. The main reason for this phenomenon is that the lidar points successfully hit on the object will decrease rapidly with the increase of measurement range. When the constraints are not sufficient, and the initial value of optimization is not precise enough, the optimization results diverge. However, if the object is in the near field of view, the estimation method based on the measured point cloud will be more accurate than that of 3D-BB inferred by the semantic detection module. We can draw this conclusion by comparing the semantic and converged geometric tracking results. In addition, the object motion trajectories of different tracking methods under sequence 0926-0056 in Fig.8 can also qualitatively verify our conclusion.

We hope to use the directly measured object point cloud to improve state estimation accuracy. Meanwhile, the tracker's robustness should be guaranteed when the object moves away from the robot. Therefore, using a fusion estimation method with semantic and geometric information seems natural. As shown in Table IV, our tracking method does not exhibit significant object trajectory drift and achieves better localization accuracy on the test sequences.
\section{Conclusions and Future Work}

In this paper, we propose a novel multi-object lidar odometry (MLO) system that aims to solve the problem of simultaneous localization, mapping, and multi-object tracking using only a lidar sensor. Specifically, we propose a fused least squares estimator using the semantic bounding box and object point cloud for object state update of multi-object modules (SGF-MOT). In the mapping module, dynamic semantic objects are detected based on the maintained absolute trajectory tracking list, and the system can achieve reliable mapping in highly dynamic scenarios. Experiments on open datasets show that, compared with the state of art MOT works, the SGF-MOT method can achieve a better balance between the accuracy and robustness of object tracking than the semantic-only and geometry-only methods. Meanwhile, compared with the representative feature-based 3D lidar SLAM methods, the MLO system with the SGF-MOT tracker can provide more stable and accurate localization in complex scenes. In the future, based on the existing MLO system, we will do an in-depth analysis and make use of variable object information to achieve a more robust environment awareness and mapping system for mobile robots.

\section*{APPENDIX}
The appendix will explain the Jacobian of  Eq. (11) and (12) in detail. In the process of derivation, we omit the superscript and subscript of variables without ambiguity. First, we use the perturbation model $ {\delta _{\bf{\theta }}} $ for the rotating part to obtain the new closest point $ {}^l{\bf{\Pi }}_{i - 1,r}^{b{\kern 1pt} {\kern 1pt} {\kern 1pt} {\kern 1pt} {\kern 1pt} {\kern 1pt}  * } $:

$$
\begin{array}{l}
{}^l{\bf{\Pi }}_{i - 1,r}^{b{\kern 1pt} {\kern 1pt} {\kern 1pt} {\kern 1pt} {\kern 1pt} {\kern 1pt}  * } =  - \left( {{\bf{I}} + {{\left[ {{\delta _{\bf{\theta }}}} \right]}_ \times }} \right){\bf{Rn}}{{\bf{t}}^T}\left( {{\bf{I}} + {{\left[ {{\delta _{\bf{\theta }}}} \right]}_ \times }} \right){\bf{Rn}}\\
{\kern 1pt} {\kern 1pt} {\kern 1pt} {\kern 1pt} {\kern 1pt} {\kern 1pt} {\kern 1pt} {\kern 1pt} {\kern 1pt} {\kern 1pt} {\kern 1pt} {\kern 1pt} {\kern 1pt} {\kern 1pt} {\kern 1pt} {\kern 1pt} {\kern 1pt} {\kern 1pt} {\kern 1pt} {\kern 1pt} {\kern 1pt} {\kern 1pt} {\kern 1pt} {\kern 1pt} {\kern 1pt} {\kern 1pt} {\kern 1pt} {\kern 1pt} {\kern 1pt} {\kern 1pt} {\kern 1pt} {\kern 1pt} {\kern 1pt} {\kern 1pt} {\kern 1pt} {\kern 1pt} {\kern 1pt} {\kern 1pt} {\kern 1pt} {\kern 1pt} {\kern 1pt} {\kern 1pt} {\kern 1pt} {\kern 1pt} {\kern 1pt} {\kern 1pt} {\kern 1pt} {\kern 1pt} {\kern 1pt} {\kern 1pt} {\kern 1pt} {\kern 1pt} {\kern 1pt} {\kern 1pt} {\kern 1pt} {\kern 1pt} {\kern 1pt} {\kern 1pt} {\kern 1pt} {\kern 1pt} {\kern 1pt} {\kern 1pt} {\kern 1pt} {\kern 1pt} {\kern 1pt} {\kern 1pt} {\kern 1pt} {\kern 1pt} {\kern 1pt} {\kern 1pt} {\kern 1pt} {\kern 1pt} {\kern 1pt} {\kern 1pt} {\kern 1pt} {\kern 1pt} {\kern 1pt} {\kern 1pt} {\kern 1pt} {\kern 1pt} {\kern 1pt} {\kern 1pt} {\kern 1pt} {\kern 1pt} {\kern 1pt} {\kern 1pt} {\kern 1pt} {\kern 1pt} {\kern 1pt} {\kern 1pt} {\kern 1pt} {\kern 1pt} {\kern 1pt} {\kern 1pt} {\kern 1pt} {\kern 1pt} {\kern 1pt} {\kern 1pt} {\kern 1pt} {\kern 1pt} {\kern 1pt} {\kern 1pt} {\kern 1pt} {\kern 1pt} {\kern 1pt} {\kern 1pt} {\kern 1pt} {\kern 1pt} {\kern 1pt} {\kern 1pt} {\kern 1pt} {\kern 1pt} {\kern 1pt} {\kern 1pt} {\kern 1pt} {\kern 1pt} {\kern 1pt} {\kern 1pt} {\kern 1pt} {\kern 1pt} {\kern 1pt} {\kern 1pt} {\kern 1pt} {\kern 1pt} {\kern 1pt} {\kern 1pt} {\kern 1pt} {\kern 1pt} {\kern 1pt} {\kern 1pt} {\kern 1pt} {\kern 1pt} {\kern 1pt} {\kern 1pt} {\kern 1pt} {\kern 1pt} {\kern 1pt} {\kern 1pt} {\kern 1pt} {\kern 1pt} {\kern 1pt} {\kern 1pt} {\kern 1pt} {\kern 1pt} {\kern 1pt} {\kern 1pt} {\kern 1pt} {\kern 1pt} {\kern 1pt}  + \left( {{\bf{I}} + {{\left[ {{\delta _{\bf{\theta }}}} \right]}_ \times }} \right){\bf{Rn}}d\\
{\kern 1pt} {\kern 1pt} {\kern 1pt} {\kern 1pt} {\kern 1pt} {\kern 1pt} {\kern 1pt} {\kern 1pt} {\kern 1pt} {\kern 1pt} {\kern 1pt} {\kern 1pt} {\kern 1pt} {\kern 1pt} {\kern 1pt} {\kern 1pt} {\kern 1pt} {\kern 1pt} {\kern 1pt} {\kern 1pt} {\kern 1pt} {\kern 1pt} {\kern 1pt} {\kern 1pt} {\kern 1pt} {\kern 1pt} {\kern 1pt} {\kern 1pt} {\kern 1pt} {\kern 1pt} {\kern 1pt}  =  - \left( {{\bf{Rn}}{{\bf{t}}^T}{\bf{Rn}} + {{\left[ {{\delta _{\bf{\theta }}}} \right]}_ \times }{\bf{Rn}}{{\bf{t}}^T}{\bf{Rn}} + {\bf{Rn}}{{\bf{t}}^T}{{\left[ {{\delta _{\bf{\theta }}}} \right]}_ \times }{\bf{Rn}}} \right)\\
{\kern 1pt} {\kern 1pt} {\kern 1pt} {\kern 1pt} {\kern 1pt} {\kern 1pt} {\kern 1pt} {\kern 1pt} {\kern 1pt} {\kern 1pt} {\kern 1pt} {\kern 1pt} {\kern 1pt} {\kern 1pt} {\kern 1pt} {\kern 1pt} {\kern 1pt} {\kern 1pt} {\kern 1pt} {\kern 1pt} {\kern 1pt} {\kern 1pt} {\kern 1pt} {\kern 1pt} {\kern 1pt} {\kern 1pt} {\kern 1pt} {\kern 1pt} {\kern 1pt} {\kern 1pt} {\kern 1pt} {\kern 1pt} {\kern 1pt} {\kern 1pt} {\kern 1pt} {\kern 1pt} {\kern 1pt} {\kern 1pt} {\kern 1pt} {\kern 1pt} {\kern 1pt} {\kern 1pt} {\kern 1pt} {\kern 1pt} {\kern 1pt} {\kern 1pt} {\kern 1pt} {\kern 1pt} {\kern 1pt} {\kern 1pt} {\kern 1pt} {\kern 1pt} {\kern 1pt} {\kern 1pt} {\kern 1pt} {\kern 1pt} {\kern 1pt} {\kern 1pt} {\kern 1pt} {\kern 1pt} {\kern 1pt} {\kern 1pt} {\kern 1pt} {\kern 1pt} {\kern 1pt} {\kern 1pt} {\kern 1pt} {\kern 1pt} {\kern 1pt} {\kern 1pt} {\kern 1pt} {\kern 1pt} {\kern 1pt} {\kern 1pt} {\kern 1pt} {\kern 1pt} {\kern 1pt} {\kern 1pt} {\kern 1pt} {\kern 1pt} {\kern 1pt} {\kern 1pt} {\kern 1pt} {\kern 1pt} {\kern 1pt} {\kern 1pt} {\kern 1pt} {\kern 1pt} {\kern 1pt} {\kern 1pt} {\kern 1pt} {\kern 1pt} {\kern 1pt} {\kern 1pt} {\kern 1pt} {\kern 1pt} {\kern 1pt} {\kern 1pt} {\kern 1pt} {\kern 1pt} {\kern 1pt} {\kern 1pt} {\kern 1pt} {\kern 1pt} {\kern 1pt} {\kern 1pt} {\kern 1pt} {\kern 1pt} {\kern 1pt} {\kern 1pt} {\kern 1pt} {\kern 1pt} {\kern 1pt} {\kern 1pt} {\kern 1pt} {\kern 1pt} {\kern 1pt} {\kern 1pt} {\kern 1pt} {\kern 1pt} {\kern 1pt} {\kern 1pt} {\kern 1pt} {\kern 1pt} {\kern 1pt} {\kern 1pt} {\kern 1pt} {\kern 1pt} {\kern 1pt} {\kern 1pt} {\kern 1pt} {\kern 1pt} {\kern 1pt} {\kern 1pt} {\kern 1pt} {\kern 1pt} {\kern 1pt} {\kern 1pt} {\kern 1pt} {\kern 1pt} {\kern 1pt} {\kern 1pt} {\kern 1pt} {\kern 1pt} {\kern 1pt} {\kern 1pt} {\kern 1pt} {\kern 1pt} {\kern 1pt} {\kern 1pt} {\kern 1pt}  + {\bf{Rn}}d + {\left[ {{\delta _{\bf{\theta }}}} \right]_ \times }{\bf{Rn}}d\\
{\kern 1pt} {\kern 1pt} {\kern 1pt} {\kern 1pt} {\kern 1pt} {\kern 1pt} {\kern 1pt} {\kern 1pt} {\kern 1pt} {\kern 1pt} {\kern 1pt} {\kern 1pt} {\kern 1pt} {\kern 1pt} {\kern 1pt} {\kern 1pt} {\kern 1pt} {\kern 1pt} {\kern 1pt} {\kern 1pt} {\kern 1pt} {\kern 1pt} {\kern 1pt} {\kern 1pt} {\kern 1pt} {\kern 1pt} {\kern 1pt} {\kern 1pt} {\kern 1pt} {\kern 1pt} {\kern 1pt}  =  - {\bf{Rn}}{{\bf{t}}^T}{\bf{Rn}} + {\bf{Rn}}d - {\left[ {{\delta _{\bf{\theta }}}} \right]_ \times }{\bf{Rn}}{{\bf{t}}^T}{\bf{Rn}}\\
{\kern 1pt} {\kern 1pt} {\kern 1pt} {\kern 1pt} {\kern 1pt} {\kern 1pt} {\kern 1pt} {\kern 1pt} {\kern 1pt} {\kern 1pt} {\kern 1pt} {\kern 1pt} {\kern 1pt} {\kern 1pt} {\kern 1pt} {\kern 1pt} {\kern 1pt} {\kern 1pt} {\kern 1pt} {\kern 1pt} {\kern 1pt} {\kern 1pt} {\kern 1pt} {\kern 1pt} {\kern 1pt} {\kern 1pt} {\kern 1pt} {\kern 1pt} {\kern 1pt} {\kern 1pt} {\kern 1pt} {\kern 1pt} {\kern 1pt} {\kern 1pt} {\kern 1pt} {\kern 1pt} {\kern 1pt} {\kern 1pt} {\kern 1pt} {\kern 1pt} {\kern 1pt} {\kern 1pt} {\kern 1pt} {\kern 1pt} {\kern 1pt} {\kern 1pt} {\kern 1pt} {\kern 1pt} {\kern 1pt} {\kern 1pt} {\kern 1pt} {\kern 1pt} {\kern 1pt} {\kern 1pt} {\kern 1pt} {\kern 1pt} {\kern 1pt} {\kern 1pt} {\kern 1pt} {\kern 1pt} {\kern 1pt} {\kern 1pt} {\kern 1pt} {\kern 1pt} {\kern 1pt} {\kern 1pt} {\kern 1pt} {\kern 1pt} {\kern 1pt} {\kern 1pt} {\kern 1pt} {\kern 1pt} {\kern 1pt} {\kern 1pt} {\kern 1pt} {\kern 1pt} {\kern 1pt} {\kern 1pt} {\kern 1pt} {\kern 1pt} {\kern 1pt} {\kern 1pt} {\kern 1pt} {\kern 1pt} {\kern 1pt} {\kern 1pt} {\kern 1pt} {\kern 1pt} {\kern 1pt} {\kern 1pt} {\kern 1pt} {\kern 1pt} {\kern 1pt} {\kern 1pt} {\kern 1pt} {\kern 1pt} {\kern 1pt} {\kern 1pt} {\kern 1pt} {\kern 1pt} {\kern 1pt} {\kern 1pt} {\kern 1pt} {\kern 1pt} {\kern 1pt} {\kern 1pt} {\kern 1pt} {\kern 1pt} {\kern 1pt} {\kern 1pt} {\kern 1pt} {\kern 1pt} {\kern 1pt} {\kern 1pt} {\kern 1pt}  - {\bf{Rn}}{{\bf{t}}^T}{\left[ {{\delta _{\bf{\theta }}}} \right]_ \times }{\bf{Rn}} + {\left[ {{\delta _{\bf{\theta }}}} \right]_ \times }{\bf{Rn}}d\\
{\kern 1pt} {\kern 1pt} {\kern 1pt} {\kern 1pt} {\kern 1pt} {\kern 1pt} {\kern 1pt} {\kern 1pt} {\kern 1pt} {\kern 1pt} {\kern 1pt} {\kern 1pt} {\kern 1pt} {\kern 1pt} {\kern 1pt} {\kern 1pt} {\kern 1pt} {\kern 1pt} {\kern 1pt} {\kern 1pt} {\kern 1pt} {\kern 1pt} {\kern 1pt} {\kern 1pt} {\kern 1pt} {\kern 1pt} {\kern 1pt} {\kern 1pt} {\kern 1pt} {\kern 1pt} {\kern 1pt}  = {}^l{\bf{\hat \Pi }}_{i - 1,r}^b + {\left[ {{\bf{Rn}}{{\bf{t}}^T}{\bf{Rn}}} \right]_ \times }{\delta _{\bf{\theta }}} + {\bf{Rn}}{{\bf{t}}^T}{\left[ {{\bf{Rn}}} \right]_ \times }{\delta _{\bf{\theta }}}\\
{\kern 1pt} {\kern 1pt} {\kern 1pt} {\kern 1pt} {\kern 1pt} {\kern 1pt} {\kern 1pt} {\kern 1pt} {\kern 1pt} {\kern 1pt} {\kern 1pt} {\kern 1pt} {\kern 1pt} {\kern 1pt} {\kern 1pt} {\kern 1pt} {\kern 1pt} {\kern 1pt} {\kern 1pt} {\kern 1pt} {\kern 1pt} {\kern 1pt} {\kern 1pt} {\kern 1pt} {\kern 1pt} {\kern 1pt} {\kern 1pt} {\kern 1pt} {\kern 1pt} {\kern 1pt} {\kern 1pt} {\kern 1pt} {\kern 1pt} {\kern 1pt} {\kern 1pt} {\kern 1pt} {\kern 1pt} {\kern 1pt} {\kern 1pt} {\kern 1pt} {\kern 1pt} {\kern 1pt} {\kern 1pt} {\kern 1pt} {\kern 1pt} {\kern 1pt} {\kern 1pt} {\kern 1pt} {\kern 1pt} {\kern 1pt} {\kern 1pt} {\kern 1pt} {\kern 1pt} {\kern 1pt} {\kern 1pt} {\kern 1pt} {\kern 1pt} {\kern 1pt} {\kern 1pt} {\kern 1pt} {\kern 1pt} {\kern 1pt} {\kern 1pt} {\kern 1pt} {\kern 1pt} {\kern 1pt} {\kern 1pt} {\kern 1pt} {\kern 1pt} {\kern 1pt} {\kern 1pt} {\kern 1pt} {\kern 1pt} {\kern 1pt} {\kern 1pt} {\kern 1pt} {\kern 1pt} {\kern 1pt} {\kern 1pt} {\kern 1pt} {\kern 1pt} {\kern 1pt} {\kern 1pt} {\kern 1pt} {\kern 1pt} {\kern 1pt} {\kern 1pt} {\kern 1pt} {\kern 1pt} {\kern 1pt} {\kern 1pt} {\kern 1pt} {\kern 1pt} {\kern 1pt} {\kern 1pt} {\kern 1pt} {\kern 1pt} {\kern 1pt} {\kern 1pt} {\kern 1pt} {\kern 1pt} {\kern 1pt} {\kern 1pt} {\kern 1pt} {\kern 1pt} {\kern 1pt} {\kern 1pt} {\kern 1pt} {\kern 1pt} {\kern 1pt} {\kern 1pt} {\kern 1pt} {\kern 1pt} {\kern 1pt} {\kern 1pt} {\kern 1pt} {\kern 1pt} {\kern 1pt} {\kern 1pt} {\kern 1pt} {\kern 1pt} {\kern 1pt} {\kern 1pt} {\kern 1pt} {\kern 1pt} {\kern 1pt} {\kern 1pt} {\kern 1pt} {\kern 1pt} {\kern 1pt} {\kern 1pt} {\kern 1pt} {\kern 1pt} {\kern 1pt} {\kern 1pt} {\kern 1pt} {\kern 1pt} {\kern 1pt} {\kern 1pt} {\kern 1pt} {\kern 1pt} {\kern 1pt} {\kern 1pt} {\kern 1pt} {\kern 1pt} {\kern 1pt} {\kern 1pt} {\kern 1pt} {\kern 1pt}  - {\bf{Rn}}d{\delta _{\bf{\theta }}}
\end{array} \eqno{(17)}
$$

Then, the Jacobian of $ {{\bf{e}}_{{b_r}}} $ with respect to the axis angle increment $ {\delta _{\bf{\theta }}} $ is:

$$
\frac{{\partial {{\bf{e}}_{{b_r}}}}}{{\partial {\delta _{\bf{\theta }}}}} =  - {\left[ {{\bf{Rn}}{{\bf{t}}^T}{\bf{Rn}}} \right]_ \times } - {\bf{Rn}}{{\bf{t}}^T}{\left[ {{\bf{Rn}}} \right]_ \times } + {\bf{Rn}}d \eqno{(18)}
$$

Similarly, we can get the new closest point $ {}^l{\bf{\Pi }}_{i - 1,r}^{b{\kern 1pt} {\kern 1pt} {\kern 1pt} {\kern 1pt} {\kern 1pt} {\kern 1pt}  * } $ by using the perturbation model ${\delta _{\bf{t}}} $ for the translation part:

$$
\begin{array}{l}
{}^l{\bf{\Pi }}_{i - 1,r}^{b{\kern 1pt} {\kern 1pt} {\kern 1pt} {\kern 1pt} {\kern 1pt} {\kern 1pt}  * } =  - {\bf{Rn}}{\left( {{\bf{t}} + {\delta _{\bf{t}}}} \right)^T}{\bf{Rn}} + {\bf{Rn}}d\\
{\kern 1pt} {\kern 1pt} {\kern 1pt} {\kern 1pt} {\kern 1pt} {\kern 1pt} {\kern 1pt} {\kern 1pt} {\kern 1pt} {\kern 1pt} {\kern 1pt} {\kern 1pt} {\kern 1pt} {\kern 1pt} {\kern 1pt} {\kern 1pt} {\kern 1pt} {\kern 1pt} {\kern 1pt} {\kern 1pt} {\kern 1pt} {\kern 1pt} {\kern 1pt} {\kern 1pt} {\kern 1pt} {\kern 1pt} {\kern 1pt} {\kern 1pt} {\kern 1pt} {\kern 1pt} {\kern 1pt}  =  - {\bf{Rn}}{{\bf{t}}^T}{\bf{Rn}} + {\bf{Rn}}d - {\bf{Rn}}{\delta _{\bf{t}}}^T{\bf{Rn}}\\
{\kern 1pt} {\kern 1pt} {\kern 1pt} {\kern 1pt} {\kern 1pt} {\kern 1pt} {\kern 1pt} {\kern 1pt} {\kern 1pt} {\kern 1pt} {\kern 1pt} {\kern 1pt} {\kern 1pt} {\kern 1pt} {\kern 1pt} {\kern 1pt} {\kern 1pt} {\kern 1pt} {\kern 1pt} {\kern 1pt} {\kern 1pt} {\kern 1pt} {\kern 1pt} {\kern 1pt} {\kern 1pt} {\kern 1pt} {\kern 1pt} {\kern 1pt} {\kern 1pt} {\kern 1pt} {\kern 1pt}  = {}^l{\bf{\hat \Pi }}_{i - 1,r}^b - {\bf{Rn}}{\left( {{\bf{Rn}}} \right)^T}{\delta _{\bf{t}}}
\end{array} \eqno{(19)}
$$

The Jacobian of $ {{\bf{e}}_{{b_r}}} $ with respect to the translation $ {\delta _{\bf{t}}} $ is:

$$
\frac{{\partial {{\bf{e}}_{{b_r}}}}}{{\partial {\delta _{\bf{t}}}}} = {\bf{Rn}}{\left( {{\bf{Rn}}} \right)^T} \eqno{(20)}
$$

\bibliographystyle{IEEEtrans}
\bibliography{mybibfile.bib}{}

\end{document}